\pdfoutput=1

\documentclass[11pt]{article}

\usepackage[]{EMNLP2022}

\usepackage{times}
\usepackage{latexsym}

\usepackage[T1]{fontenc}

\usepackage[utf8]{inputenc}

\usepackage{microtype}

\usepackage{inconsolata}

\usepackage{algorithm}
\usepackage[noend]{algpseudocode}
\usepackage{amsmath}
\usepackage{amssymb}
\usepackage{array}
\usepackage{arydshln}
\usepackage{bm}
\usepackage{booktabs}
\usepackage{color}
\usepackage{colortbl}
\usepackage{fancyvrb}
\usepackage[T1]{fontenc}
\usepackage{graphicx}
\usepackage{multirow}
\usepackage{pifont}
\usepackage{stfloats}
\usepackage{xcolor}

\definecolor{right}{RGB}{0,128,96}
\definecolor{wrong}{RGB}{192,0,32}
\newcommand{\Right}[1]{\textcolor{right}{#1}}

\DeclareMathOperator*{\argmax}{arg\,max}

\newcommand{\Ds}{\mathcal{D}}
\newcommand{\E}{\mathbb{E}}
\newcommand{\eps}{\varepsilon}
\newcommand{\Pqa}[1]{P_{\text{QA}}(#1)}
\newcommand{\Sqa}[1]{S_{\text{QA}}(#1)}
\newcommand{\pqa}[1]{p_{\text{QA}}(#1)}

\newcommand{\sgn}{\text{sgn}}

\definecolor{difcolor}{RGB}{0,0,192}
\newcommand{\dif}[1]{\textcolor{difcolor}{#1}}
\renewcommand{\dif}[1]{#1}

\newcommand{\methodname}{\textsc{Rainier}}

\title{\includegraphics[scale=0.09]{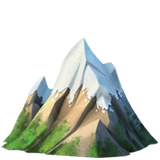} \methodname: Reinforced Knowledge Introspector \\ for Commonsense Question Answering}

\newcommand{\aspace}{\hspace{1em}}
\newcommand{\uw}{$^{\heartsuit}$}
\newcommand{\aitwo}{$^{\spadesuit}$}

\author{%
  Jiacheng Liu\uw \aspace
  Skyler Hallinan\uw \aspace
  Ximing Lu\uw \aitwo \aspace
  Pengfei He\uw \aspace \\
  \textbf{Sean Welleck}\uw \aitwo \aspace
  \textbf{Hannaneh Hajishirzi}\uw \aitwo \aspace
  \textbf{Yejin Choi}\uw \aitwo \aspace \\
  \uw{}Paul G. Allen School of Computer Science \& Engineering, University of Washington \\
  \aitwo{}Allen Institute for Artificial Intelligence \\
  \texttt{liujc@cs.washington.edu}
}

\begin{document}

\maketitle

\begin{abstract}
Knowledge underpins reasoning. Recent research demonstrates that when relevant knowledge is provided as additional context to commonsense question answering (QA), it can substantially enhance the performance even on top of state-of-the-art. The fundamental challenge is where and how to find such knowledge that is high quality and on point with respect to the question; knowledge retrieved from knowledge bases are incomplete and knowledge generated from language models are inconsistent.

We present \methodname{}\footnote{Code, model and knowledge-extended datasets are available at \url{http://github.com/liujch1998/rainier}}, or \textit{Reinforced Knowledge Introspector}, that learns to generate contextually relevant knowledge in response to given questions. Our approach starts by imitating knowledge generated by GPT-3, then learns to generate its own knowledge via reinforcement learning where rewards are shaped based on the increased performance on the resulting question answering. \methodname{} demonstrates substantial and consistent performance gains when tested over 9 different commonsense benchmarks: including 5 datasets that are seen during model training, as well as 4 datasets that are kept unseen. Our work is the first to report that knowledge generated by models that are orders of magnitude smaller than GPT-3, even without direct supervision on the knowledge itself, can exceed the quality of commonsense knowledge elicited from GPT-3. 
%
\end{abstract}

\section{Introduction}
\label{sec:introduction}

\begin{figure}[t]
\centering
\includegraphics[width=\linewidth]{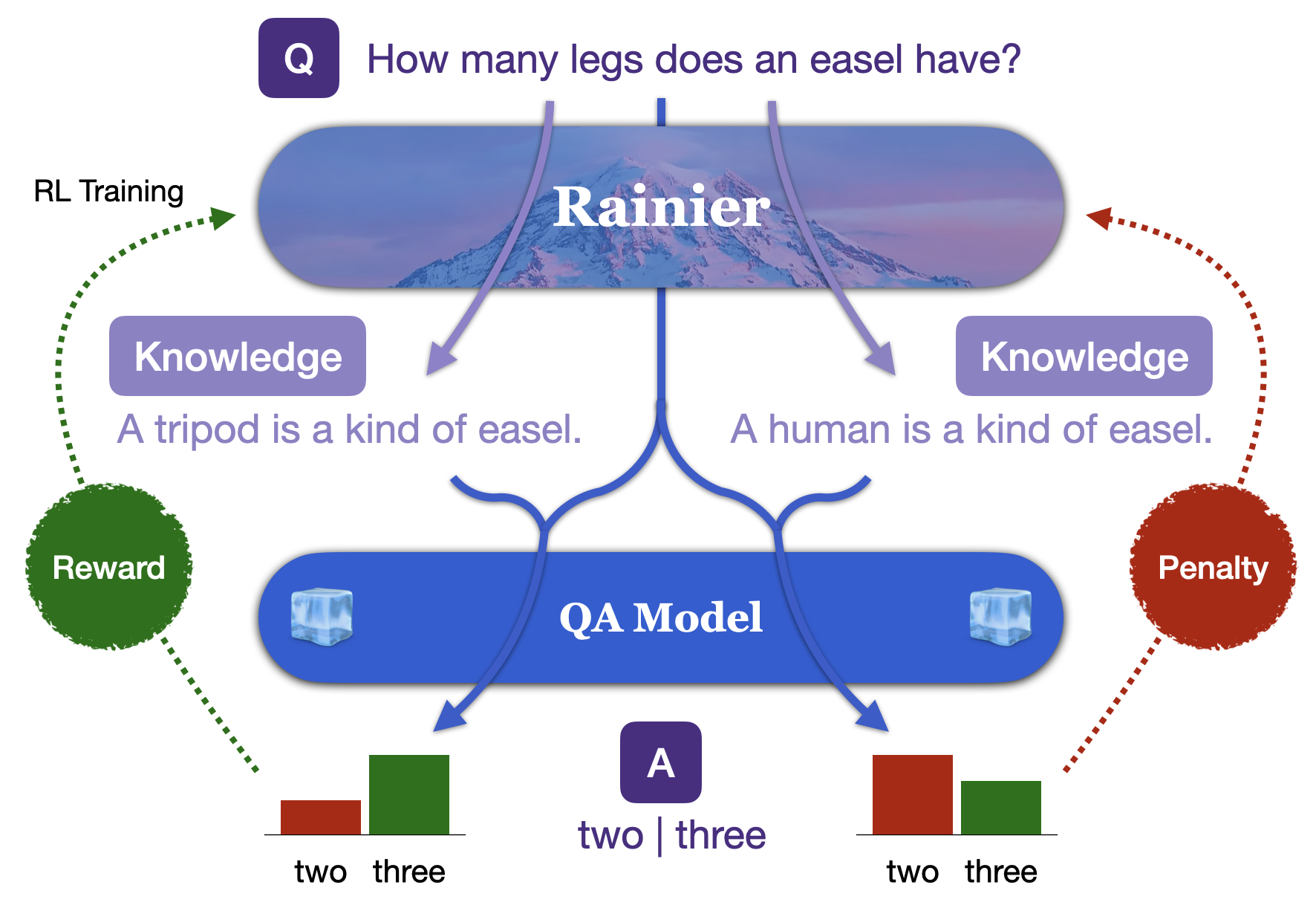}
\caption{
    \methodname{} can introspect for commonsense knowledge that underpin the reasoning process, and is trained via reinforcement learning, where the reward is derived from the effectiveness of knowledge when prompting a frozen, generic QA model.
}
\label{fig:rainier}
\vspace{-16pt}
\end{figure}


\dif{
Commonsense is a significant challenge for modern NLP models, due to the obscurity of underlying knowledge that grounds the reasoning process.
While humans are generally able to \textit{introspect} the underlying reasons for their conclusion \citep{Mercier2017TheEO},
neural models
lack the capability to verbalize the premises leading to their prediction.
This hinders models' performance and robustness on commonsense tasks, and makes it difficult to inspect their point of failure.
Recent research has demonstrated that relevant knowledge can provide useful context for approaching commonsense tasks.
Yet these methods either retrieve from in-domain knowledge bases \cite{mitra2019additional, chang2021incorporating} that do not have good coverage over commonsense, or generate knowledge from neural models \citep{shwartz2020unsupervised, gu2021dream, liu2021generated}, which often need domain-specific engineering and very large models (e.g. GPT-3 \citep{brown2020language}).
It is still an open challenge to systematically find high-quality knowledge.

In this work, we \dif{use} a novel, reinforcement-learning-based method to develop \methodname{}, a generative neural model that can introspect the underlying knowledge for reasoning about commonsense questions. As illustrated in Fig \ref{fig:rainier}, \methodname{} is trained to  generate knowledge that are both fluent natural language statements, and  useful \textit{prompts} that optimize the performance of a generic question answering (QA) model. 
Our model (1) demonstrates strong generalization to unseen benchmarks \dif{without additional engineering effort (e.g. finetuning)},
(2) \dif{produces commonsense knowledge of high quality and diversity,}
and (3) \dif{is substantially smaller in size compared to GPT-3, the best knowledge source reported so far.}

}

\dif{To train \methodname{}, we optimize knowledge introspection for the resulting QA, instead of direct supervision, because there are usually no gold knowledge labels on commonsense datasets.}
\dif{In order to ensure that our model learns to generate generically useful knowledge for a broad range of QA models, we train only \methodname{}, the knowledge introspector, without finetuning the QA model.}
Since our desired knowledge are sequences of discrete, non-differentiable word tokens, we adapt a reinforcement learning algorithm, Proximal Policy Optimization (PPO) \citep{schulman2017proximal, ouyang2022training}, to optimize the knowledge introspector.
Specifically, the reward is defined as the effect of \methodname{}-generated knowledge on the QA model's prediction.
We train \methodname{} in a multi-task setting on 8 commonsense QA datasets -- encompassing general, scientific, physical, and social commonsense -- to equip the model with better generalization to unseen benchmarks.

Experiments show that \methodname{} substantially improves the performance of QA models on 9 commonsense benchmarks (5 datasets seen during training and 4 unseen datasets), and gives larger and more consistent gains than a few-shot GPT-3 \citep{liu2021generated} despite being 16x smaller in parameter size.
It also boosts the performance on top of those QA models that it is not trained against, indicating that it generates generically useful knowledge instead of merely hacking into the reward given by a single QA model.
Knowledge generated by \methodname{} can even boost a QA model that is 4x larger than it, showing the promise of using model-generated knowledge as a complement to model scaling in making progress in commonsense reasoning.
Our analyses show that the knowledge generated by \methodname{} are of high quality, and are diverse in terms of domain (e.g. scientific, social), relation expressed (e.g. part of, member of, purpose), and syntactic property (e.g. negation, comparison).
The effect of these knowledge on the QA model also aligns well with human judgments.
The success of \methodname{} shows that moderately-sized models can serve as source of high-quality and useful commonsense knowledge that facilitates reasoning.
We publicly release the code, the trained \methodname{} model, and the commonsense datasets extended with knowledge generated by \methodname{}.

\section{Method}
\label{sec:method}




\paragraph{Problem Overview.}
We focus on the tasks of multiple-choice commonsense QA, consisting of instances of format $x = (q, A, a^*)$, where $q$ is the question, $A$ is the set of candidate answers, and $a^* \in A$ is the correct answer.
For full contextualization, we append candidate answers $A$ to the question $q$ to form the input to the QA model as follows:
\begin{align*}
q = \texttt{\small \{question\} (A) \{choice\_A\} (B) \{choice\_B\} ...}
\end{align*}
Common approaches only train supervised QA models.
As a complement,
we train a separate model, which we refer to as \methodname{}, that can introspect question-specific knowledges
that are useful to \textit{prompt} a fixed QA model.
\methodname{} is a sequence-to-sequence language model, $p_K(k | q; \theta)$, and we expect it to generate knowledge statements ($k$'s) in response to the given question ($q$).
However, the challenge is that we have no gold knowledge labels as supervision.

\begin{figure}[t]
\begin{algorithm}[H]
\footnotesize
\caption{
    Training \methodname{}
}
\textbf{Input} initial policy model $\theta_0$, initial value model $\phi_0$, pre-trained QA model $\psi_{\text{QA}}$ \\
 \hspace*{2mm} $\Ds_{\text{imit}} \gets$ Get silver knowledge on $\Ds_{\text{seen}}$ from GPT-3. \\
  \hspace*{2mm} $\theta_{\text{imit}} \gets$ Optimize $\theta_0$ with Eqn~\ref{eqn:imit-loss} from $\Ds_{\text{imit}}$. \algorithmiccomment{Section \ref{sec:method_imit}} \\
   \hspace*{2mm} $\theta_{\methodname{}} \gets$ \Call{ReinforcedLearning}{$\Ds_{\text{seen}}$, $\theta_{\text{imit}}$, $\phi_0$, $\psi_{\text{QA}}$} \\
\hspace*{2mm} \phantom{...} \algorithmiccomment{Section \ref{sec:method_rl}} 
\begin{algorithmic}
\Procedure{ReinforcedLearning}{$\Ds_{\text{seen}}$, $\theta$, $\phi$, $\psi_{\text{QA}}$}
    \State $\theta_{\text{old}} \gets \theta$, $\phi_{\text{old}} \gets \phi$
    \For{iterations = 1, 2, \ldots}
        \State Sample a minibatch from $\Ds_{\text{seen}}$.
        \For{step = 1, 2, \ldots, $s$}
            \State Compute $\mathcal{L}_{\text{PPO}}$ on the minibatch with Eqn~\ref{eqn:ppo-loss}.
            \State Optimize $\theta$ and $\phi$ with $\mathcal{L}_{\text{PPO}}$ for one step.
        \EndFor
        \State $\theta_{\text{old}} \gets \theta$, $\phi_{\text{old}} \gets \phi$
    \EndFor
    \State \Return $\theta$
\EndProcedure
\end{algorithmic}
\textbf{Output} $\theta_{\methodname{}}$
\label{alg:ppo}
\end{algorithm}
\end{figure}

\paragraph{Training.}
Since we do not have gold knowledge to train \methodname{}, we obtain this model by finetuning a pretrained language model in two stages: (I) imitation learning, and then (II) reinforcement learning.
In Stage I (\S\ref{sec:method_imit}), we get \dif{\textit{silver}}
knowledge labels on some datasets from GPT-3, and teach our model to imitate this knowledge-generating GPT-3.
This equips our model with the basic functionality of knowledge generation.
In Stage II (\S\ref{sec:method_rl}),
we use reinforcement learning to continue training the model obtained in Stage I to make the generated knowledge more useful while staying fluent and meaningful. Specially, we set the reward to be the effect of the generated knowledge on the prediction made by a fixed, generic QA model.
We obtain silver knowledge and train \methodname{} on the union of multiple QA datasets (which are considered \textit{seen} during training), i.e. $\Ds_{\text{seen}} = \bigcup_{d=1}^{\Delta_{\text{seen}}}{\Ds_{d}}$, where $\Ds_d = \{ (q_j, A_j, a^*_j) \}_{j=1}^{|\Ds_d|}$.
The generic QA model we use may or may not have been trained on these seen datasets.
The complete training process is outlined in Algorithm~\ref{alg:ppo}.

\paragraph{Inference.}
The effectiveness of \methodname{} is evaluated against a set of \textit{unseen} QA datasets, $\Ds_{\text{unseen}}$, in addition to the seen datasets.
Note that \methodname{} is not trained on any unseen datasets, which means we neither get silver knowledge, nor do imitation learning or reinforcement learning on them.
The generic QA model we use was not trained on any unseen datasets as well.
We discuss details of inference in \S\ref{sec:method_inf}.


\subsection{Training Stage I: Imitation Learning}
\label{sec:method_imit}

In Stage I, we train \methodname{} so that it generates fluent and meaningful natural language statements that resemble knowledge.
\dif{There is no large-scale commonsense QA dataset labeled with high-quality knowledge, but GPT-3 has been shown as a good generator for relevant knowledge \citep{liu2021generated}.}
Therefore, we get silver knowledge from GPT-3 on our seen datasets. 
Following \citet{liu2021generated}, we elicit question-related knowledge by prompting GPT-3 with a task-specific set of few-shot demonstrations (See \S\ref{sec:prompts} for details on the prompts), and decoding $M$ knowledge for each question:
\begin{align*}
K(q) &= \big\{ k_m : k_m \sim p_G(k \mid \text{prompt}(\text{task}(q)), q) \big\}, 
\end{align*}
where $p_G(\cdot | \cdot)$ denotes GPT-3 with nucleus sampling where $p = 0.5$ \citep{holtzman2019curious}.
This yields a silver dataset of question-knowledge pairs:
\begin{align}
\Ds_{\text{imit}} &= \Big\{ (q, k) : (q, A, a^*) \in \Ds_{\text{seen}}, k \in K(q) \Big\},
\label{eqn:Dimit-construction}
\end{align}

We then train \methodname{}, starting from a pretrained sequence-to-sequence language model, on this silver dataset with standard supervised loss: 
\begin{align}
\mathcal{L}^{\text{train}}(\theta) &\propto \sum_{(q, k) \in \Ds_{\text{imit}}^{\text{train}}}{-\log{p_K(k | q; \theta)}}.
\label{eqn:imit-loss}
\end{align}
The parameterization of the resulting model is denoted as $\theta_{\text{imit}}$.

\subsection{Training Stage II: Reinforcement Learning}
\label{sec:method_rl}

As we will see in the empirical results, the imitation model obtained in Stage I does not provide the most beneficial knowledge.
Therefore, in Stage II, we continue optimizing \methodname{} to generate  knowledge that \textit{best} prompts the QA model, by directly maximizing the reward given by this QA model.

\paragraph{Knowledge generation as reinforcement learning.}
Since knowledge statements ($k$'s) are discrete and thus non-differentiable, we adopt a reinforcement learning approach, and consider knowledge generation as a sequential decision making process over the natural language vocabulary space.
We consider the generation of knowledge statement $k$ with  $T$ tokens as an episode of length $T$.
At step $t \in [1, T]$, the state $s_t = (q, k_{<t})$ is the combination of the question and the knowledge decoded up to the $(t-1)$-th token; the action $a_t = k_t$ would be the $t$-th token to decode.
The \methodname{} model, $p_K(k_t | q, k_{<t}; \theta)$, \dif{is}
the \textit{policy model} that we optimize.
We define a reward function $r(x, k)$ that characterizes the effect of the knowledge on the QA model's prediction, and discuss the definition of this reward function in \S\ref{sec:reward}.


To ensure that the generated knowledge stay fluent and meaningful, we would like the learned policy model not to move too far from the initial imitation model.
Therefore, we add to the reward an (approximate) KL penalty between the learned policy and the initial policy \citep{ouyang2022training},
\begin{align*}
R(x, k) = r(x, k) - \beta \log{\frac{p_K(k | q; \theta)}{p_K(k | q; \theta_{\text{imit}})}}.
\end{align*}
Since this reward is computed based on the full knowledge statement, we assign it to the last step of the episode.
Non-terminal steps are assigned zero rewards.
Formally,
\begin{align*}
r_T &= R(x, k) & (\text{where } T = |k| \text{ and } k_T = \text{[EOS]}); \\
r_t &= 0 & (\text{where } 1 \le t < T).
\end{align*}


We employ Proximal Policy Optimization\footnote{\dif{We choose PPO because it has shown successful results in other NLP tasks \citep{nakano2021webgpt, stiennon2020learning}. Our earlier experiments with REINFORCE did not show promising results.}} (PPO) \citep{schulman2017proximal} as our reinforcement learning algorithm, and adapt from the implementation of PPO in \citet{ouyang2022training}.
Aside from the policy model, PPO additionally uses a \textit{value model} (parameterized by $\phi$) to estimate the value function for states with incomplete decoded text, i.e. $V(s_t; \phi)$ for any $t$.
PPO minimizes a joint loss,
\begin{align}
\mathcal{L}_{\text{PPO}}(\theta, \phi) &= \mathcal{L}_{\text{Policy}}(\theta) + \alpha \cdot \mathcal{L}_{\text{Value}}(\phi),
\label{eqn:ppo-loss}
\end{align}
where $\mathcal{L}_{\text{Policy}}(\theta)$ is the loss on the policy model, $\mathcal{L}_{\text{Value}}(\phi)$ is the loss on the value model, and $\alpha$ is a hyperparameter.

\paragraph{Policy loss.}
\dif{
To obtain the policy loss, we first compute the \textit{truncated estimated advantage function},
\begin{align*}
& \hat{A}_t = \sum_{t'=t}^{T-1}{(\gamma \lambda)^{t'-t} \delta_{t'}}, \\
& \text{where} \quad \delta_{t'} = r_{t'} + \gamma V(s_{t'+1}; \phi) - V(s_{t'}; \phi),
\end{align*}
where the value functions \dif{$V(\cdot)$} are estimated by the value model.
}
PPO then maximizes the empirical expectation of a so-called \textit{clipped surrogate objective} term,
\begin{align*}
&\text{cso}(\hat{A}_t, \nu_t(\theta), \eps) = \\ &\qquad \min \big( \nu_t(\theta) \hat{A}_t, \text{clip}(\nu_t(\theta), 1 - \eps, 1 + \eps) \hat{A}_t \big),
\end{align*}
where $\nu_t(\theta) = \frac{p_K(k_t | q; \theta)}{p_K(k_t | q; \theta_{\text{old}})}$ is the ratio between the current policy $\theta$ and a lagging policy $\theta_{\text{old}}$.
The lagging policy is updated to the current policy under a fixed interval of $s$ training steps, and is kept fixed otherwise.
We adapt this to our use case, and define the policy loss as
\begin{align*}
\mathcal{L}_{\text{Policy}}(\theta) &= -\hat{\E} \Big[ \text{cso}(\hat{A}_t, \nu_t(\theta), \eps) \Big]
\end{align*}
where the expectation is taken over all instances in the training data ($x \sim \Ds_{\text{seen}}^{\text{train}}$), the distribution of model-generated knowledge as determined by the current policy conditioning on the instance's question ($k \sim p_K(k | q; \theta)$), and all tokens in the knowledge statement ($t \in [1, |k|]$).

\paragraph{Value loss.}
The value model is trained with MSE loss with respect to the target value, $V^{\text{targ}}_t$, which in turn is estimated with a lagging value model $\phi_{\text{old}}$:
\begin{align*}
& \mathcal{L}_{\text{Value}}(\phi) = \hat{\E} \Big[ \big( V(s_t; \phi) - V^{\text{targ}}_t \big)^2 \Big], \\
& \text{where} \quad V^{\text{targ}}_t = \sum_{t'=t}^{T-1}{\gamma^{t'-t} r_{t'}} + \gamma^{T-t} V(s_T; \phi_{\text{old}}).
\end{align*}


\subsubsection{Reward Shaping}
\label{sec:reward}

\dif{
We define the reward function in reinforcement learning as the quantified effect of \methodname{}'s knowledge on the QA model's prediction.
Suppose we already have a reasonably good QA model, which assigns a probability score $\Pqa{a|q}$ to any candidate answer $a \in A$.
Since we will use a sequence-to-sequence language model (i.e. UnifiedQA \citep{khashabi2020unifiedqa}) as the QA model, we define
\begin{align*}
\Pqa{a | q} = \frac{\exp{\Sqa{a | q}}}{\sum_{a' \in A}{\exp{\Sqa{a' | q}}}},
\end{align*}
where
\begin{align*}
\Sqa{a | q} = \frac{1}{|a|} \sum_{i=1}^{|a|}{-\log{\pqa{a_i | q, a_{<i}; \psi_{\text{QA}}}}},
\end{align*}
where $\pqa{a_i | q, a_{<i}; \psi_{\text{QA}}}$ is the language modeling score received by $a_i$, the $i$-th token of $a$.
The naive prediction would be the candidate answer that gets the highest $\Pqa{a|q}$ (or equivalently, the highest $\Sqa{a|q}$): $\hat{a} = \argmax_{a \in A} \Pqa{a | q}$.
}

\dif{
We aim at maximizing $\Pqa{a^* | q \circ k}$, the probability score received by the correct answer when the QA model is prompted with the knowledge $k$ generated by \methodname{}, and $\circ$ denotes text concatenation.
One naive definition of reward function may be
\begin{align*}
r(x, k) = \Pqa{a^* | q \circ k} - \Pqa{a^* | q}.
\end{align*}
However, this reward only captures the absolute change of score, but not whether the model prediction is changed or not.
To remedy for this, we define the reward function as
\begin{align*}
& r(x, k) = \frac{1}{2} \Big[ \\
& \quad \tanh \big( \Sqa{a^* | q \circ k} - \max_{\substack{a' \in A, \\ a' \ne a^*}} \Sqa{a' | q \circ k} \big) \\ & \quad - \tanh \big( \Sqa{a^* | q} - \max_{\substack{a' \in A, \\ a' \ne a^*}} \Sqa{a' | q} \big) \Big].
\end{align*}
Intuitively, this function would give a reward of near $+1$ if the naive prediction is incorrect (i.e. $\Sqa{a^* | q} < \max_{a' \in A, a' \ne a^*} \Sqa{a' | q}$), while the knowledge-prompted prediction is correct (i.e. $\Sqa{a^* | q \circ k} > \max_{a' \in A, a' \ne a^*} \Sqa{a' | q \circ k}$).
Similarly, the reward would be near $-1$ if the naive prediction is correct but the knowledge-prompted prediction is incorrect.
The hyperbolic tangent serves as a smoothed sign function, and provides a soft interpolation between the two polarity of reward values by taking into account the margin of the correct answer.
}

We also experiment with some alternative definitions of the reward function. See \autoref{tab:ablation_rewards}.

\paragraph{Reward normalization.}
To stabilize training, we apply an affine transformation on the rewards so that initially they are normalized.
Before starting Stage II training, we use the imitation model to generate a knowledge statement for each training instance, and estimate the population mean and standard deviation of rewards:
\begin{align}
\mathcal{R}_{\text{init}} &= \big\{ r(x, k) : x \in \Ds_{\text{seen}}^{\text{train}}, k \sim p_K(\cdot | q; \theta_{\text{imit}}) \big\}, \nonumber \\
\mu_0 &= \mu(\mathcal{R}_{\text{init}}), \sigma_0 = \sigma(\mathcal{R}_{\text{init}}).
\label{eqn:reward-norm-compute}
\end{align}
In Stage II training, each reward is normalized as:
\begin{align}
r(x, k) &\leftarrow \frac{r(x, k) - \mu_0}{\sigma_0}.
\label{eqn:reward-norm-apply}
\end{align}

\subsection{Inference: Knowledge Prompting and Aggregation}
\label{sec:method_inf}

Following \citet{liu2021generated}, at inference time we use \methodname{} to generate multiple knowledge per question, and \textit{prompt} the QA model by individually concatenating each knowledge to the question.
The knowledge are generated by \methodname{} with nucleus sampling where $p = 0.5$ \citep{holtzman2019curious},
\begin{align*}
K(q) &= \{\eps\} \cup \big\{ k_m : k_m \sim p_K^{p=0.5}(k \mid q; \theta), \\
& \qquad \qquad \qquad \quad m = 1 \hdots M \big\},
\end{align*}
where $M$ is the number of knowledge per question, and $\eps$ denotes empty string.
We collect a set of outputs for prompting with each knowledge.
The final prediction is the candidate answer that receives maximum confidence,
\begin{align*}
\hat{a} &= \argmax_{a \in A} \max_{k \in K(q)} \Pqa{a | q \circ k},
\end{align*}
and the prediction is supported by a single knowledge -- the \textit{selected knowledge},
\begin{align*}
\hat{k} &= \argmax_{k \in K(q)} \max_{a \in A} \Pqa{a | q \circ k}.
\end{align*}

\paragraph{Training time model selection.}
In Stage II training, we only generate one knowledge per question for the validation set.\footnote{This is for efficiency purposes. We use greedy decoding here because it is more stable than nucleus sampling when generating only one knowledge per question.}
Predictions are made using the same knowledge prompting method as above, and the model checkpoint with the maximal accuracy on the union of all validation sets is selected.
\section{Experiments}
\label{sec:experiments}

\paragraph{Seen datasets.}
For both imitation learning and reinforcement learning, we use the 8 multiple-choice datasets that $\text{UnifiedQA}_{\text{v2}}$ \citep{khashabi2022unifiedqa} uses for training:
OpenBookQA \citep{mihaylov2018can}, ARC \citep{clark2018think}, AI2Science \citep{clark2018think}, CommonsenseQA \citep{talmor2018commonsenseqa}, QASC \citep{khot2020qasc}, PhysicalIQA \citep{bisk2020piqa}, SocialIQA \citep{sap2019socialiqa}, and Winogrande \citep{sakaguchi2021winogrande}.\footnote{We exclude MCTest and RACE because most questions in these reading comprehension datasets are too long to fit into our model's input.}

\paragraph{Unseen datasets.}
We additionally evaluate our method on the following 4 multiple-choice QA datasets that our model was \textit{not} trained on:
NumerSense \citep{lin2020birds}, RiddleSense \citep{lin2021riddlesense}, QuaRTz \citep{tafjord2019quartz}, and HellaSwag \citep{zellers2019hellaswag}.

\paragraph{Models.}
For Stage I training, we get silver knowledge from the GPT-3-Curie (13B) model \citep{brown2020language}.
The knowledge introspector is initialized with T5-large \citep{raffel2019exploring}, which has 0.77B parameters.
For Stage II training, we initialize the value model with T5-large, and replace the language modeling head with a value regression head, which is initialized from scratch; we use UnifiedQA-large (UQA-large) \citep{khashabi2020unifiedqa} as the QA model that provides reward, which means the text concatenation function is defined as $q \circ k =$ \texttt{\{q\} \textbackslash{}n \{k\}}.
We use the same question formatting as UnifiedQA.
See \autoref{tab:hypers} for hyperparameters.


\paragraph{Baselines.}
We mainly report performance improvements over the vanilla QA baseline (i.e. direct inference with the UnifiedQA-large model and without prompting \methodname{}-generated knowledge).
We also consider using knowledge from:
\begin{itemize}
\item Few-shot GPT-3 \citep{liu2021generated}, where knowledge statements are elicited from the GPT-3-Curie (13B) model -- under the same prompts used for getting silver knowledge in Stage I training (\S\ref{sec:method_imit}), and the same hyperparameter setting for decoding ($M = 10$ knowledge per question, with nucleus sampling where $p = 0.5$).
\item Self-talk \citep{shwartz2020unsupervised}, where we generate $M = 10$ knowledge per question with GPT-3-Curie and a variety of templates.
\item DREAM \citep{gu2021dream}, where we generate $M = 10$ scene elaborations per question with the DREAM (11B) model.
\end{itemize}
See \S\ref{sec:baselines} for more details on these baselines.
We do not compare with chain-of-thought prompting \citep{wei2022chain} because it relies on emergent behaviors that does not exist in the scale that we experiment with.

\section{Results}
\label{sec:results}

\subsection{Main Results}
\label{sec:result_main}


\begin{table*}[t]
\setlength{\tabcolsep}{3pt}
\centering
\resizebox{\textwidth}{!}{%
\begin{tabular}{l cc cc cc cc cc cc}
\toprule
\textbf{Dataset} $\rightarrow$ & \multicolumn{2}{c}{CSQA} & \multicolumn{2}{c}{QASC} & \multicolumn{2}{c}{PIQA} & \multicolumn{2}{c}{SIQA} & \multicolumn{2}{c}{WG} & \multicolumn{2}{c}{\textbf{Avg.}} \\
\textbf{Method} $\downarrow$ & \small{dev} & \small{test} & \small{dev} & \small{test} & \small{dev} & \small{test} & \small{dev} & \small{test} & \small{dev} & \small{test} & \small{dev} & \small{test} \\
\midrule
UQA-large (0.77B) & 61.43 & 53.00 & 43.09 & 45.65 & 63.66 & 65.50 & 53.84 & 57.21 & 53.35 & 54.67 & 55.07 & 55.21 \\
\quad + Few-shot GPT-3-Curie (13B) & 66.34 & -- & 53.24 & -- & 64.25 & -- & \textbf{58.29} & -- & 55.56 & -- & 59.54 & -- \\
\quad + Self-talk GPT-3-Curie (13B) & 63.31 & -- & 49.89 & -- & 65.23 & -- & 51.89 & -- & 52.96 & -- & 56.66 & -- \\
\quad + DREAM (11B) & 64.54 & -- & 49.46 & -- & 64.74 & -- & 51.59 & -- & 56.12 & -- & 57.29 & -- \\
\midrule
\quad \textbf{+ \methodname{}-large (0.77B) [ours]} & \textbf{67.24} & \textbf{60.18} & \textbf{54.97} & \textbf{54.13} & \textbf{65.67} & \textbf{67.09} & 57.01 & \textbf{59.01} & \textbf{56.91} & \textbf{57.39} & \textbf{60.36} & \textbf{59.56} \\
\bottomrule
\end{tabular}
}%
\vspace{-4pt}
\caption{
    Results on \textbf{seen} datasets.
    All experiments use UnifiedQA-large as the QA model, and optionally uses knowledge from one of the knowledge generation models.
    Skipped baselines are marked with ``--''.
}
\label{tab:results_id}
\vspace{16pt}
\resizebox{0.85\textwidth}{!}{%
\begin{tabular}{l cc cc cc cc cc}
\toprule
\textbf{Dataset} $\rightarrow$ & \multicolumn{2}{c}{NS} & \multicolumn{2}{c}{RS} & \multicolumn{2}{c}{QuaRTz} & \multicolumn{2}{c}{HS} & \multicolumn{2}{c}{\textbf{Avg.}} \\
\textbf{Method} $\downarrow$ & \small{dev} & \small{test-all} & \small{dev} & \small{test} & \small{dev} & \small{test} & \small{dev} & \small{test} & \small{dev} & \small{test} \\
\midrule
UQA-large (0.77B) & 26.50 & 19.61 & 28.11 & 38.34 & 68.75 & 67.60 & 35.00 & 34.30 & 39.59 & 41.85 \\
\quad + Few-shot GPT-3-Curie (13B) & 38.00 & -- & 35.65 & -- & 69.01 & -- & 37.33 & -- & 45.00 & -- \\
\midrule
\quad \textbf{+ \methodname{}-large (0.77B) [ours]} & 30.00 & 21.81 & 30.07 & 41.22 & 70.31 & 68.24 & 35.73 & 34.80 & 41.53 & 43.76 \\
\bottomrule
\end{tabular}
}%
\vspace{-4pt}
\caption{
    Results on \textbf{unseen} datasets.
}
\label{tab:results_ood}
\end{table*}
\paragraph{Performance on seen datasets.}
\autoref{tab:results_id} shows the performance of \methodname{}-enhanced QA model on the seen datasets.
On average, our method achieves more than 5\% improvement over directly applying the QA model.
The knowledge generated by \methodname{} improves performance on five benchmarks: CommonsenseQA, QASC, PhysicalIQA, SocialIQA, and Winogrande, with the greatest improvement on CommonsenseQA (+6\%) and QASC (+12\%).
As shown in \autoref{tab:results_id_nope}, there is no performance gain on OpenBookQA, ARC, and AI2Science.
We conjecture that this is because the QA model, UnifiedQA, is already trained on these three datasets, thus setting a strong baseline.

\paragraph{Comparison with other models.}
Compared to \methodname{}, other knowledge generation models, including few-shot GPT-3, Self-talk, and DREAM, provide generally weaker improvements over the vanilla QA baseline.
In particular, \methodname{} outperforms GPT-3-based models
while being 16x smaller in parameter size (0.77B vs. 13B).

\paragraph{Performance on unseen datasets.}
\autoref{tab:results_ood} shows that \methodname{}'s knowledge substantially improves performance over the vanilla QA baseline on the four unseen datasets, demonstrating its generalization capability.

\begin{figure*}
\begin{minipage}{.48\linewidth}
\centering
\includegraphics[width=\linewidth]{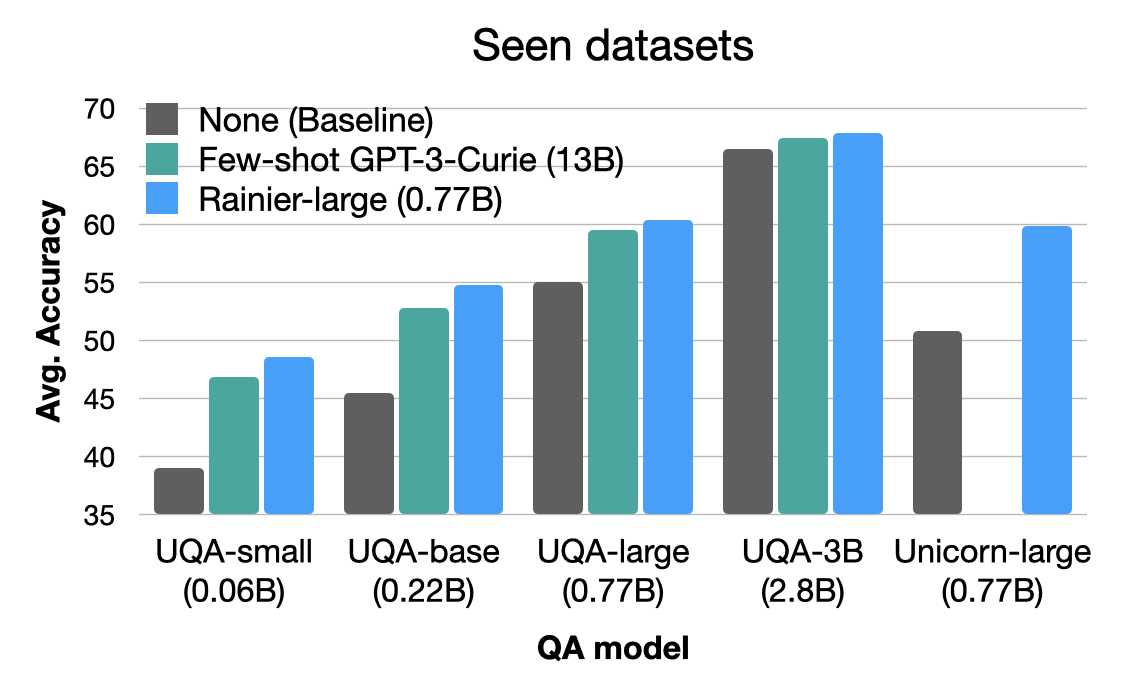}
\end{minipage}
\hfill
\begin{minipage}{.48\linewidth}
\centering
\includegraphics[width=\linewidth]{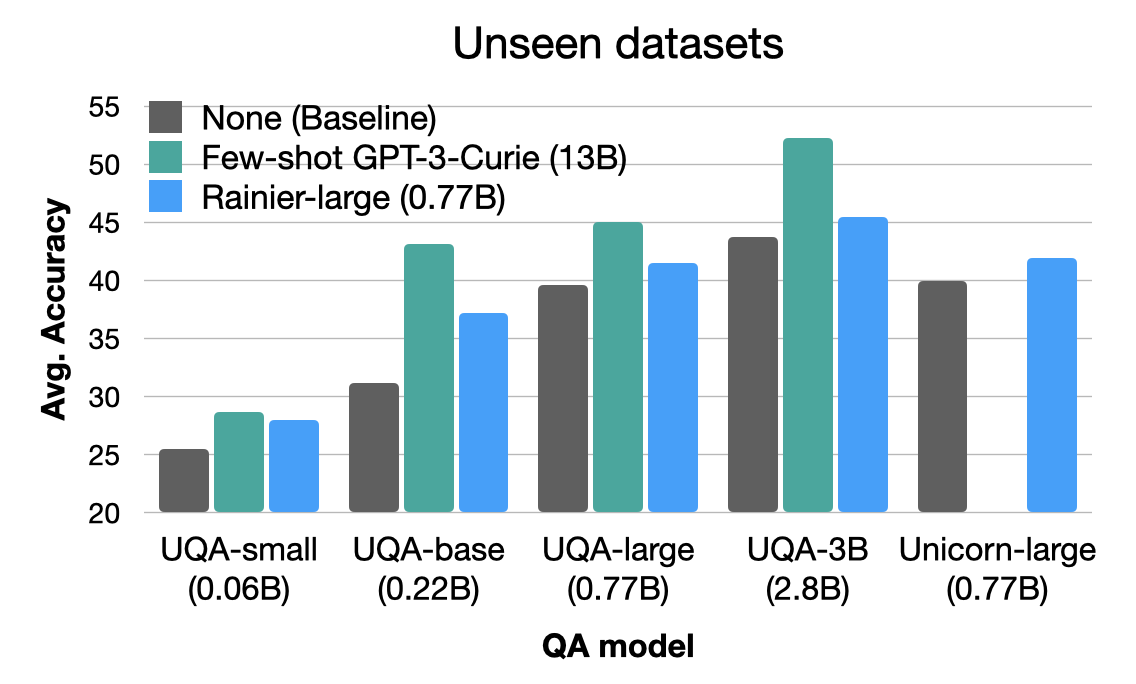}
\end{minipage}
\vspace{-4pt}
\caption{
    Effectiveness of \methodname{}-generated knowledge on different QA models.
    Average accuracy on dev sets is reported.
    (Note: results of few-shot GPT-3-Curie on Unicorn-large is missing.)
}
\label{fig:results_qa}
\end{figure*}
\paragraph{Choice of QA model for evaluation.}
To verify that our \methodname{} model is not hacking into the rewards provided by the QA model we use during training, we evaluate the effect of \methodname{}'s knowledge on different QA models.
We choose three other UnifiedQA models with different sizes, as well as a different model known as Unicorn \citep{lourie2021unicorn}.
Results are shown in \autoref{fig:results_qa}.
\methodname{} consistently gives performance gains on top of all QA models, indicating that its knowledge are generally useful information rather than mere artifacts of model-specific reward hacking.
We even observe performance gains with a QA model that is 4x as large as \methodname{}, which means generating and prompting relevant knowledge can be a technique complementary to model scaling, and can be done meaningfully with smaller models.
Finally, we see the largest improvement when the QA model itself has weak, but non-trivial, performance (UnifiedQA-small for seen datasets, and UnifiedQA-base for unseen datasets).

\subsection{Ablations}
\label{sec:ablations}


\begin{table*}[t]
\begin{minipage}[t]{.30 \linewidth}
\setlength{\tabcolsep}{4pt}
\centering
\resizebox{1.0 \textwidth}{!}{%
\begin{tabular}{l cc}
\toprule
\textbf{QA Model} $\rightarrow$ & \multicolumn{2}{c}{UQA-large} \\
\textbf{Knowledge Gen.} $\downarrow$ & seen & unseen \\
\midrule
None & 55.07 & 39.59 \\
\midrule
\methodname{}-large & \textbf{60.36} & \textbf{41.53} \\
\quad -- Stage I & 53.68 & 36.83 \\
\quad -- Stage II & 57.00 & 40.70 \\
\quad -- Stage I -- Stage II & 53.29 & 36.72 \\
\bottomrule
\end{tabular}
}%
\vspace{-4pt}
\caption{
    Ablations on the importance of both training stages.
}
\label{tab:ablation_stages}
\end{minipage}
\hfill
\begin{minipage}[t]{.68 \linewidth}
\setlength{\tabcolsep}{4pt}
\centering
\resizebox{1.0 \textwidth}{!}{%
\begin{tabular}{ll cc}
\toprule
\textbf{QA Model} $\rightarrow$ &  & \multicolumn{2}{c}{UQA-large} \\
\textbf{Reward Func.} $\downarrow$ & \textbf{Definition:} $r(x, k) = \hdots$ & seen & unseen \\
\midrule
\methodname{}'s & $\frac{1}{2} \Big[ \tanh \big( \Sqa{a^* | q \circ k} - \max_{a' \in A, a' \ne a^*} \Sqa{a' | q \circ k} \big)$ \\ & $\qquad - \tanh \big( \Sqa{a^* | q} - \max_{a' \in A, a' \ne a^*} \Sqa{a' | q} \big) \Big]$ & 60.36 & \textbf{41.53} \\
\midrule
Prob only & $\Pqa{a^* | q \circ k}$ & 59.11 & 40.61 \\ 
Prob diff & $\Pqa{a^* | q \circ k} - \Pqa{a^* | q}$ & \textbf{60.69} & 40.91 \\ 
Score diff & $\Sqa{a^* | q \circ k} - \Sqa{a^* | q}$ & 58.26 & 39.86 \\ 
Hard activation & $\frac{1}{2} \Big[ \sgn \big( \Sqa{a^* | q \circ k} - \max_{a' \in A, a' \ne a^*} \Sqa{a' | q \circ k} \big)$ \\ & $\qquad - \sgn \big( \Sqa{a^* | q} - \max_{a' \in A, a' \ne a^*} \Sqa{a' | q} \big) \Big]$ & 58.32 & 41.16 \\ 
\bottomrule
\end{tabular}
}%
\vspace{-4pt}
\caption{
    Ablations on the choice of reward function.
}
\label{tab:ablation_rewards}
\end{minipage}
\vspace{-8pt}
\end{table*}

\paragraph{Stage I and Stage II training.}
We experimented with omitting the Stage I (imitation) and/or Stage II (reinforcement) from the training pipeline.
Results are shown in \autoref{tab:ablation_stages}.
Without Stage I training, \methodname{} does not improve the performance of the QA model \dif{(regardless of whether it is trained with Stage II or not)}, showing the indispensability of equipping the model with the basic functionality of knowledge generation.
On the other hand, a model trained solely with Stage I gives weaker improvements than the fully trained \methodname{}, stressing the importance of Stage II training as well.

\paragraph{Reward function.}
\autoref{tab:ablation_rewards} shows the results for knowledge introspectors trained with different reward functions.
Our reward shaping gives the best performance on unseen datasets, as well as one of the top performance on seen datasets.
\dif{While the naive \textit{prob diff} reward function gives slightly better performance on seen datasets, our reward shaping results in better generalization.}

\subsection{Analysis}
\label{sec:analysis}

\begin{figure*}
\begin{minipage}{.22\linewidth}
\centering
\includegraphics[width=\linewidth]{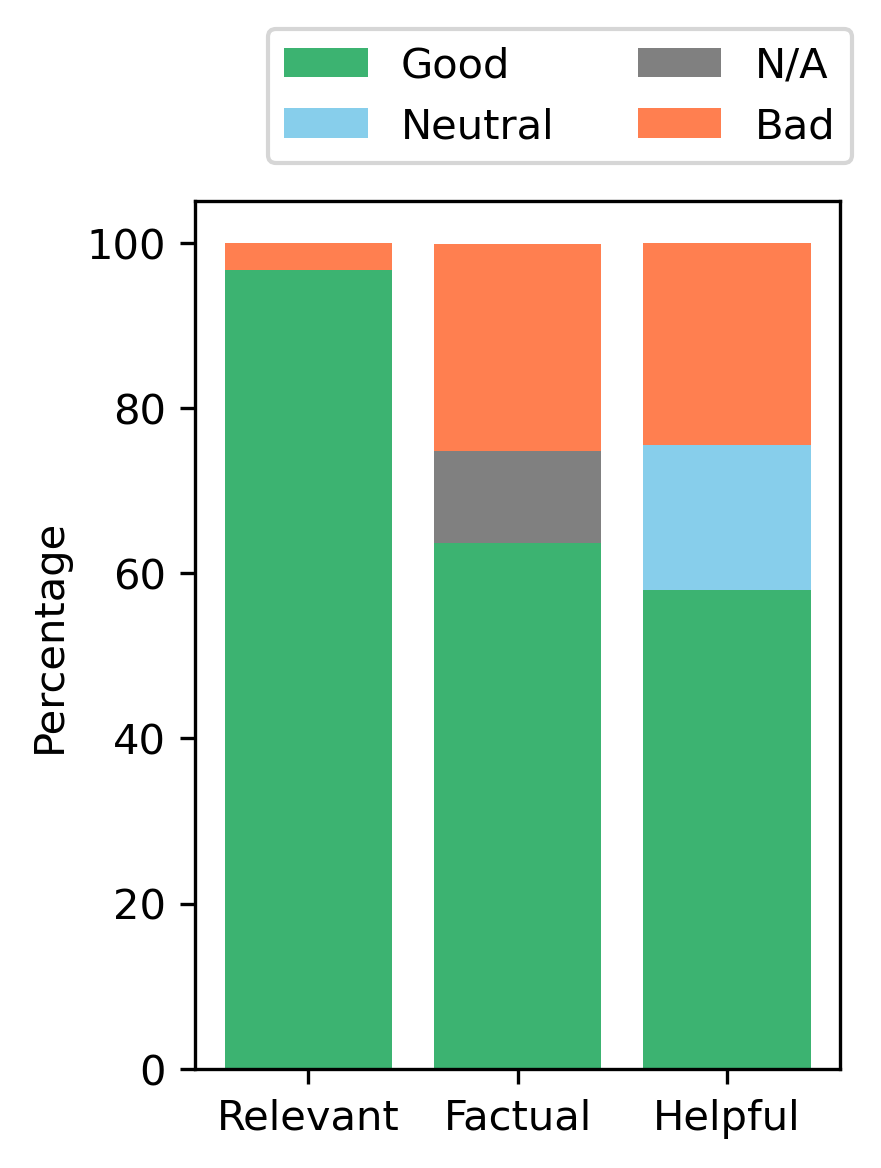}
\end{minipage}
\hfill
\begin{minipage}{.48\linewidth}
\centering
\includegraphics[width=\linewidth]{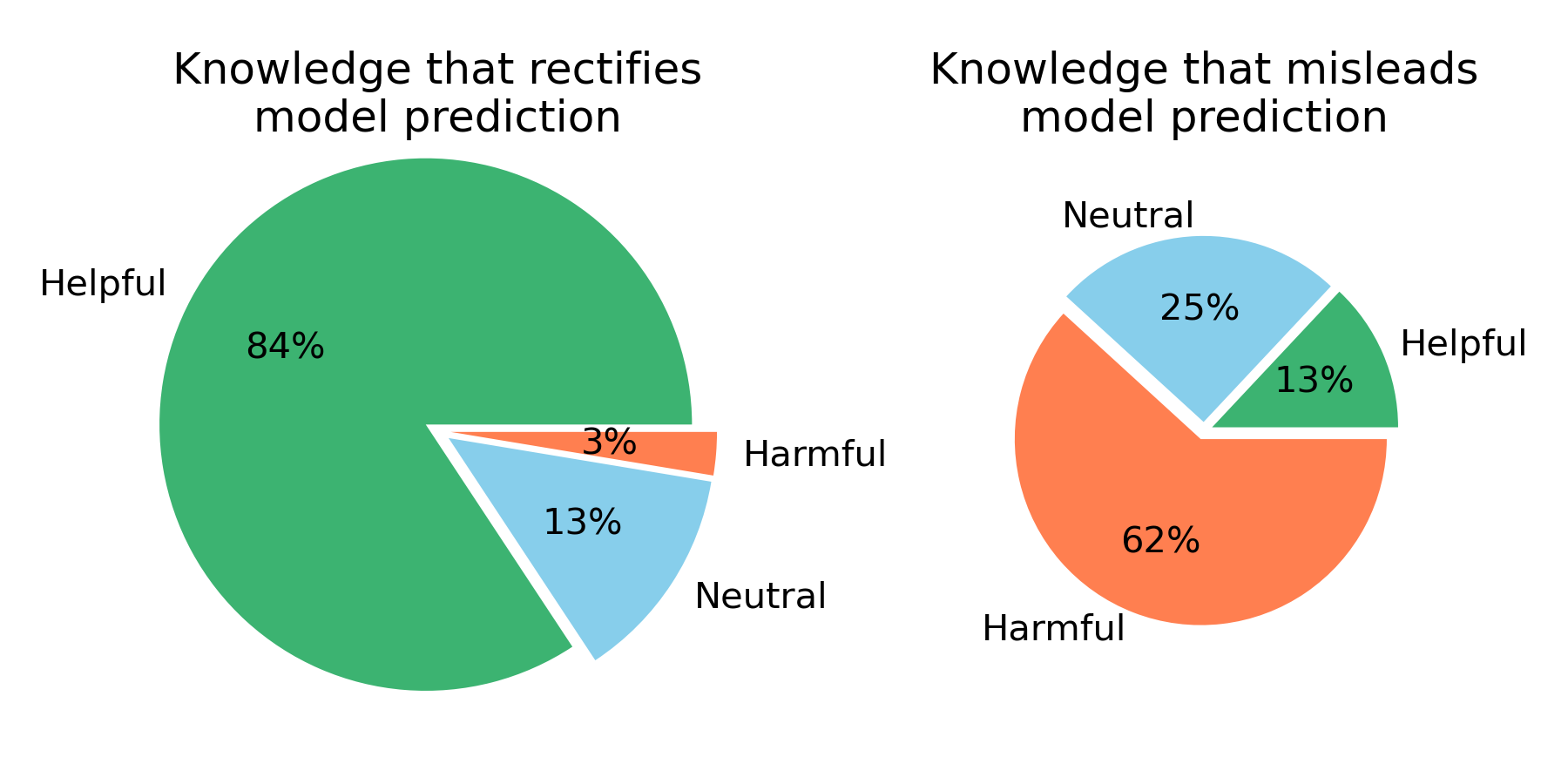}
\end{minipage}
\hfill
\begin{minipage}{.28\linewidth}
\centering
\includegraphics[width=\linewidth]{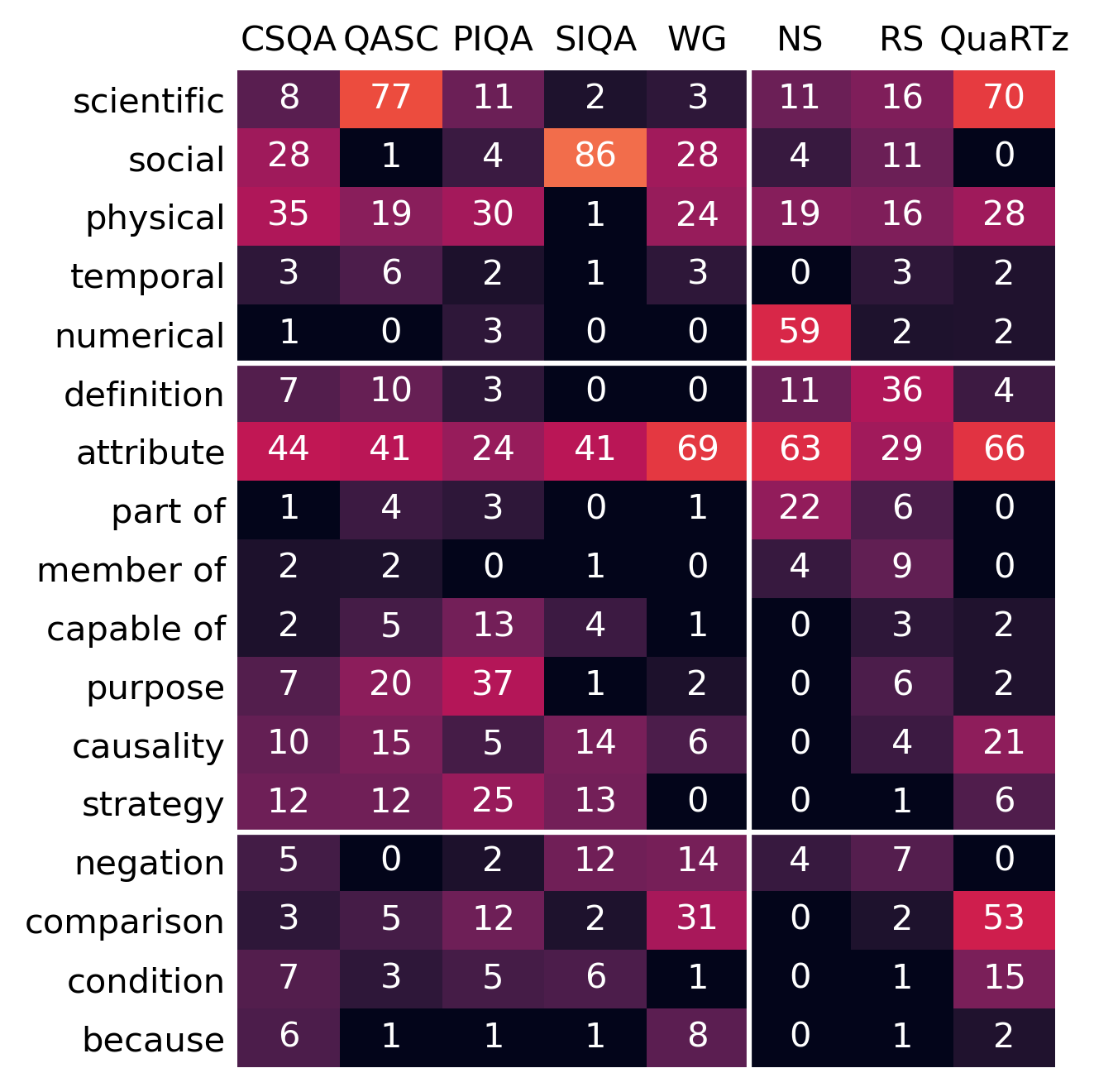}
\end{minipage}
\caption{
    Human analysis of \methodname{}-generated knowledge.
    \textbf{Left:} Percentage of good knowledge in each quality aspect.
    \textbf{Mid:} Agreement between human and machine on helpfulness of \textit{selected knowledge}.
    \textbf{Right:} Percentage of \methodname{}-generated knowledge categorized by domain, expressed relation, and syntax. The percentages do not add up to 100\% because some knowledge have none of these characteristics, while some others may have multiple.
}
\label{fig:human}
\vspace{-8pt}
\end{figure*}

\dif{
To get a deeper understanding of the behavior and capability of \methodname{}, we manually analyzed the generated knowledge along several \textbf{quality} and \textbf{diversity} aspects.
We asked three NLP experts to annotate the \textit{selected knowledge} (\S\ref{sec:method_inf}) for up to 100 questions per dataset among the validation sets of 8 benchmarks (5 seen, 3 unseen; see \autoref{fig:human}).
It was hidden from the annotators whether the knowledge rectifies or misleads QA model's prediction, so potential bias is eliminated.
}

\paragraph{Quality.}
First, we follow \citet{liu2021generated} by annotating the quality aspects -- \textit{relevance}, \textit{factuality}, and \textit{helpfulness} -- of each knowledge with respect to the question.
We find that \methodname{}-generated knowledge are overwhelmingly related to the respective questions.
64\% are factually correct, 25\% are factually incorrect, and the remaining 11\% have undetermined factuality due to various reasons (e.g. ambiguity, cultural sensitivity).
58\% are seen by human as being helpful for reasoning about the question, whereas 24\% are seen as harmful.

\dif{
In our annotations, there are 420 knowledge that \textit{rectify} UnifiedQA-large's predictions (i.e. flipping from wrong to right), and 246 knowledge that \textit{mislead} the predictions (i.e. flipping from right to wrong).
Among the rectifying knowledge, 84\% are deemed helpful by human; and among the misleading knowledge, 62\% are deemed harmful.
These results have similar trends as \citet{liu2021generated}, and show that \methodname{}'s knowledge are of high quality and interpretability in helping QA models.
}

\paragraph{Diversity.}
Additionally, we analyze the \textbf{diversity} aspects by annotating each knowledge with the \textit{domain}(s) it belongs to (e.g. scientific, social), the \textit{relation}(s) it expresses (e.g. attribute, capable of), and its \textit{syntactic} property(s) (e.g. negation, comparison).
See \autoref{fig:human} for complete list of options under each aspect.
The knowledge's domain distribution is strongly tied to the domain of the benchmark (e.g. scientific for QASC and QuaRTz, social for SocialIQA and Winogrande, numerical for NumerSense).
The domain aspect is more diverse for benchmarks that test general commonsense, like CommonsenseQA and RiddleSense.
For the relation aspect, there are many knowledge that express an ``attribute'' relation, while other relations are also substantially represented.
As for syntax, a good proportion of the knowledge contain structures like comparison and negation.
Therefore, \methodname{}'s knowledge have good syntactic and semantic diversity while being able to adapt to the domain.

\subsection{Qualitative Examples}
\label{sec:qual}

We show some examples of good knowledge generated by \methodname{} in \autoref{tab:qual}.


\section{Related Work}
\label{sec:related}

\paragraph{Explicit reasoning for commonsense QA.}
Commonsense question answering poses a significant challenge to modern neural models.
To improve performance and interpretability, many work have proposed to do explicit reasoning for tasks in this area, that is, to verbalize the intermediate text artifacts that facilitate the reasoning process.
\citet{rajani2019explain} and \citet{latcinnik2020explaining} use supervised learning to train models to generate text explanations, while \citet{gu2021dream} and \citet{bansal2021cose} use similar training regimes to obtain models that can generate scene elaborations and paths through a structured knowledge graph, respectively.
\citet{shwartz2020unsupervised} and \citet{paranjape2021prompting} prompt pretrained models with pre-defined templates to generate question clarifications or contrastive explanations, which are in turn used to prompt the inference model.
The above approaches all \dif{pose, implicitly or explicitly, certain constraints (e.g. domain, relation, syntax) on the model-generated text.}
In contrast, \citet{wei2022chain} elicits full chain-of-reasoning from language models with in-context learning; \citet{liu2021generated} uses few-shot demonstrations to elicit flexible, relevant knowledge statements from a language model, and \citet{wang2022elaboration} distills this capability into smaller models using supervised learning.
These methods provide more flexibility on the knowledge, yet they rely on accessing very large language models (e.g. GPT-3).
Aside from methods that make reasoning explicit in a linear chain manner, another set of work produce recursive structures of reasoning, through either backward chaining \citep{dalvi2022towards, jung2022maieutic} or forward chaining \citep{bostrom2022natural}.
\dif{Our work contributes to this line of research, yet we depart from prior work by presenting the first approach that \textit{learns} to generate relevant knowledge without requiring human-labeled gold knowledge.}

\paragraph{Reinforcement learning for NLP.}
Recently, reinforcement learning methods have been adopted for NLP tasks like question answering \citep{nakano2021webgpt}, summarization \citep{stiennon2020learning, paulus2017deep}, machine translation \citep{shen2015minimum, wu2016google}, grounded text generation \citep{ammanabrolu2020motivate, ammanabrolu2022aligning}, controlled text generation \citep{lu2022quark}, and prompt generation \citep{guo2021text, deng2022rlprompt}.
Our application of reinforcement learning on knowledge introspection is novel.
The idea of reinforcement learning with model-provided feedback has been previously explored in \citet{guo2021text}, \citet{ammanabrolu2020motivate}, and \citet{lu2022quark}.
The PPO algorithm has been previously employed to optimize rewards learned from human feedback \citep{nakano2021webgpt, stiennon2020learning}.
In contrast, we use PPO to optimize reward purely derived from the decision-making neural models.


\section{Conclusion}
\label{sec:conclusion}

\begin{table*}
\setlength{\tabcolsep}{3pt}
\footnotesize
\centering
\begin{tabular}{cp{280pt} p{32pt}p{40pt}p{40pt}}
\toprule
\textbf{Task} & Question / \textbf{Knowledge} & \textbf{Domain} & \textbf{Relation} & \textbf{Syntax} \\
\midrule
\multirow{2}{*}{CSQA} & What would vinyl be an odd thing to replace? (A) pants (B) record albums (C) record store (D) cheese \Right{(E) wallpaper} \\
& \qquad \textbf{Vinyl is a type of plastic.} & scientific & member of & -- \\
\midrule
\multirow{2}{*}{QASC} & Some pelycosaurs gave rise to reptile ancestral to (A) lamphreys (B) angiosperm \Right{(C) mammals} (D) paramecium (E) animals (F) protozoa (G) arachnids (H) backbones & \quad \newline \quad \newline scientific \\
& \qquad \textbf{Reptiles are the ancestors of all mammals.} & temporal & attribute & -- \\
\midrule
\multirow{2}{*}{SIQA} & Sydney rubbed Addison's head because she had a horrible headache. What will happen to Sydney? (A) drift to sleep \Right{(B) receive thanks} (C) be reprimanded \\
& \qquad \textbf{A good deed will be rewarded.} & social & -- & -- \\
\midrule
\multirow{2}{*}{WG} & Adam always spent all of the free time watching Tv unlike Hunter who volunteered, due to \_ being lazy. \Right{(A) Adam} (B) Hunter \\
& \qquad \textbf{Hunter is more active than Adam.} & social & attribute & comparison \\
\midrule
\multirow{2}{*}{RS} & Causes bad breath and frightens blood-suckers (A) tuna (B) iron (C) trash \Right{(D) garlic} (E) pubs \\
& \qquad \textbf{Garlic is a strong-smelling food.} & -- & attribute & -- \\
\midrule
\multirow{2}{*}{QuaRTz} & If the mass of an object gets bigger what will happen to the amount of matter contained within it? \Right{(A) gets bigger} (B) gets smaller & \quad \newline scientific \\
& \quad \textbf{The mass of an object is proportional to the amount of matter it contains.} & physical & -- & -- \\
\bottomrule
\end{tabular}
\caption{
    Examples of good knowledge generated by \methodname{}.
    Each of these knowledge rectifies UnifiedQA-large's prediction, and is labeled by the annotator as relevant, factual, and helpful.
}
\label{tab:qual}
\vspace{-8pt}
\end{table*}

\dif{
We introduced \methodname{}, a neural model that can introspect for relevant knowledge on a broad range of commonsense question answering tasks.
\methodname{} is trained with a novel adaption of reinforcement learning, and does not need gold knowledge labels that are difficult to obtain.
Knowledge generated by \methodname{} can serve as useful prompts that improves the performance of QA models on both seen and unseen benchmarks, and outperform knowledge elicited from a few-shot GPT-3 which is 16x bigger.
\methodname{} generates knowledge in the form of natural language statements that are fluent, meaningful, high-quality, and diverse in terms of domain and relation; furthermore, the effect of these knowledge on the QA model is found to align well with human judgments.
}


\section*{Limitations}
\label{sec:limitations}

\dif{
Despite the positive effect of our knowledge introspector \methodname{} on commonsense QA tasks,
its performance on non-commonsense applications is unknown and thus requires further investigation.
Even for commonsense applications, there is still a large gap between model performance and human performance, so the resulting model is not ready for real-world applications.
}
There is also a limit on the length of knowledge it generates in our experimental setting, and it has not been tested on generating long and coherent text.
Furthermore, in some cases it may generate knowledge that express inappropriate social values (\autoref{tab:qual_social}), are culture-specific (\autoref{tab:qual_culture}), or contain ethical risks (\autoref{tab:qual_ethics}).
See \S\ref{sec:analysis_more} for examples.
Extra care should be taken when applying our model in production environments, especially when making critical decisions or exposing its generated contents directly to human end users.

\section*{Acknowledgements}
%
This work was funded in part by the DARPA MCS program through NIWC Pacific (N66001-19-2-4031), NSF IIS-2044660, and ONR N00014-18-1-2826.
We thank OpenAI for offering access to the GPT-3 API.

We would like to thank Prithviraj Ammanabrolu, Alisa Liu and Weijia Shi for the discussion and feedback on early drafts of the paper.
We also thank the anonymous reviewers for their valuable feedback.

\bibliographystyle{acl_natbib}
\bibliography{custom}

\begin{thebibliography}{44}
\expandafter\ifx\csname natexlab\endcsname\relax\def\natexlab#1{#1}\fi

\bibitem[{Ammanabrolu et~al.(2022)Ammanabrolu, Jiang, Sap, Hajishirzi, and
  Choi}]{ammanabrolu2022aligning}
Prithviraj Ammanabrolu, Liwei Jiang, Maarten Sap, Hannaneh Hajishirzi, and
  Yejin Choi. 2022.
\newblock \href {https://arxiv.org/abs/2205.01975} {Aligning to social norms
  and values in interactive narratives}.
\newblock \emph{arXiv preprint arXiv:2205.01975}.

\bibitem[{Ammanabrolu et~al.(2021)Ammanabrolu, Urbanek, Li, Szlam,
  Rockt{\"a}schel, and Weston}]{ammanabrolu2020motivate}
Prithviraj Ammanabrolu, Jack Urbanek, Margaret Li, Arthur Szlam, Tim
  Rockt{\"a}schel, and Jason Weston. 2021.
\newblock \href {https://doi.org/10.18653/v1/2021.naacl-main.64} {How to
  motivate your dragon: Teaching goal-driven agents to speak and act in fantasy
  worlds}.
\newblock In \emph{Proceedings of the 2021 Conference of the North American
  Chapter of the Association for Computational Linguistics: Human Language
  Technologies}, pages 807--833, Online. Association for Computational
  Linguistics.

\bibitem[{Bansal et~al.(2021)Bansal, Aggarwal, Bhatia, Kaur, and
  Krishnamurthy}]{bansal2021cose}
Rachit Bansal, Milan Aggarwal, Sumit Bhatia, Jivat~Neet Kaur, and Balaji
  Krishnamurthy. 2021.
\newblock Cose-co: Text conditioned generative commonsense contextualizer.

\bibitem[{Bisk et~al.(2020)Bisk, Zellers, LeBras, Gao, and Choi}]{bisk2020piqa}
Yonatan Bisk, Rowan Zellers, Ronan LeBras, Jianfeng Gao, and Yejin Choi. 2020.
\newblock \href {https://aaai.org/ojs/index.php/AAAI/article/view/6239}
  {{PIQA:} reasoning about physical commonsense in natural language}.
\newblock In \emph{The Thirty-Fourth {AAAI} Conference on Artificial
  Intelligence, {AAAI} 2020, The Thirty-Second Innovative Applications of
  Artificial Intelligence Conference, {IAAI} 2020, The Tenth {AAAI} Symposium
  on Educational Advances in Artificial Intelligence, {EAAI} 2020, New York,
  NY, USA, February 7-12, 2020}, pages 7432--7439. {AAAI} Press.

\bibitem[{Bostrom et~al.(2022)Bostrom, Sprague, Chaudhuri, and
  Durrett}]{bostrom2022natural}
Kaj Bostrom, Zayne Sprague, Swarat Chaudhuri, and Greg Durrett. 2022.
\newblock \href {https://arxiv.org/abs/2201.06028} {Natural language deduction
  through search over statement compositions}.
\newblock \emph{arXiv preprint arXiv:2201.06028}.

\bibitem[{Brown et~al.(2020)Brown, Mann, Ryder, Subbiah, Kaplan, Dhariwal,
  Neelakantan, Shyam, Sastry, Askell, Agarwal, Herbert{-}Voss, Krueger,
  Henighan, Child, Ramesh, Ziegler, Wu, Winter, Hesse, Chen, Sigler, Litwin,
  Gray, Chess, Clark, Berner, McCandlish, Radford, Sutskever, and
  Amodei}]{brown2020language}
Tom~B. Brown, Benjamin Mann, Nick Ryder, Melanie Subbiah, Jared Kaplan,
  Prafulla Dhariwal, Arvind Neelakantan, Pranav Shyam, Girish Sastry, Amanda
  Askell, Sandhini Agarwal, Ariel Herbert{-}Voss, Gretchen Krueger, Tom
  Henighan, Rewon Child, Aditya Ramesh, Daniel~M. Ziegler, Jeffrey Wu, Clemens
  Winter, Christopher Hesse, Mark Chen, Eric Sigler, Mateusz Litwin, Scott
  Gray, Benjamin Chess, Jack Clark, Christopher Berner, Sam McCandlish, Alec
  Radford, Ilya Sutskever, and Dario Amodei. 2020.
\newblock \href
  {https://proceedings.neurips.cc/paper/2020/hash/1457c0d6bfcb4967418bfb8ac142f64a-Abstract.html}
  {Language models are few-shot learners}.
\newblock In \emph{Advances in Neural Information Processing Systems 33: Annual
  Conference on Neural Information Processing Systems 2020, NeurIPS 2020,
  December 6-12, 2020, virtual}.

\bibitem[{Chang et~al.(2020)Chang, Liu, Gopalakrishnan, Hedayatnia, Zhou, and
  Hakkani-Tur}]{chang2021incorporating}
Ting-Yun Chang, Yang Liu, Karthik Gopalakrishnan, Behnam Hedayatnia, Pei Zhou,
  and Dilek Hakkani-Tur. 2020.
\newblock \href {https://doi.org/10.18653/v1/2020.deelio-1.9} {Incorporating
  commonsense knowledge graph in pretrained models for social commonsense
  tasks}.
\newblock In \emph{Proceedings of Deep Learning Inside Out (DeeLIO): The First
  Workshop on Knowledge Extraction and Integration for Deep Learning
  Architectures}, pages 74--79, Online. Association for Computational
  Linguistics.

\bibitem[{Clark et~al.(2018)Clark, Cowhey, Etzioni, Khot, Sabharwal, Schoenick,
  and Tafjord}]{clark2018think}
Peter Clark, Isaac Cowhey, Oren Etzioni, Tushar Khot, Ashish Sabharwal, Carissa
  Schoenick, and Oyvind Tafjord. 2018.
\newblock \href {https://arxiv.org/abs/1803.05457} {Think you have solved
  question answering? try arc, the ai2 reasoning challenge}.
\newblock \emph{arXiv preprint arXiv:1803.05457}.

\bibitem[{Dalvi et~al.(2022)Dalvi, Tafjord, and Clark}]{dalvi2022towards}
Bhavana Dalvi, Oyvind Tafjord, and Peter Clark. 2022.
\newblock \href {https://arxiv.org/abs/2204.13074} {Towards teachable reasoning
  systems}.
\newblock \emph{arXiv preprint arXiv:2204.13074}.

\bibitem[{Deng et~al.(2022)Deng, Wang, Hsieh, Wang, Guo, Shu, Song, Xing, and
  Hu}]{deng2022rlprompt}
Mingkai Deng, Jianyu Wang, Cheng-Ping Hsieh, Yihan Wang, Han Guo, Tianmin Shu,
  Meng Song, Eric~P Xing, and Zhiting Hu. 2022.
\newblock \href {https://arxiv.org/abs/2205.12548} {Rlprompt: Optimizing
  discrete text prompts with reinforcement learning}.
\newblock \emph{arXiv preprint arXiv:2205.12548}.

\bibitem[{Gu et~al.(2022)Gu, Dalvi, and Clark}]{gu2021dream}
Yuling Gu, Bhavana Dalvi, and Peter Clark. 2022.
\newblock \href {https://doi.org/10.18653/v1/2022.naacl-main.82} {{DREAM}:
  Improving situational {QA} by first elaborating the situation}.
\newblock In \emph{Proceedings of the 2022 Conference of the North American
  Chapter of the Association for Computational Linguistics: Human Language
  Technologies}, pages 1115--1127, Seattle, United States. Association for
  Computational Linguistics.

\bibitem[{Guo et~al.(2021)Guo, Tan, Liu, Xing, and Hu}]{guo2021text}
Han Guo, Bowen Tan, Zhengzhong Liu, Eric~P Xing, and Zhiting Hu. 2021.
\newblock \href {https://arxiv.org/abs/2106.07704} {Text generation with
  efficient (soft) q-learning}.
\newblock \emph{arXiv preprint arXiv:2106.07704}.

\bibitem[{Holtzman et~al.(2020)Holtzman, Buys, Du, Forbes, and
  Choi}]{holtzman2019curious}
Ari Holtzman, Jan Buys, Li~Du, Maxwell Forbes, and Yejin Choi. 2020.
\newblock \href {https://openreview.net/forum?id=rygGQyrFvH} {The curious case
  of neural text degeneration}.
\newblock In \emph{8th International Conference on Learning Representations,
  {ICLR} 2020, Addis Ababa, Ethiopia, April 26-30, 2020}. OpenReview.net.

\bibitem[{Jung et~al.(2022)Jung, Qin, Welleck, Brahman, Bhagavatula, Bras, and
  Choi}]{jung2022maieutic}
Jaehun Jung, Lianhui Qin, Sean Welleck, Faeze Brahman, Chandra Bhagavatula,
  Ronan~Le Bras, and Yejin Choi. 2022.
\newblock \href {https://arxiv.org/abs/2205.11822} {Maieutic prompting:
  Logically consistent reasoning with recursive explanations}.
\newblock \emph{arXiv preprint arXiv:2205.11822}.

\bibitem[{Khashabi et~al.(2022)Khashabi, Kordi, and
  Hajishirzi}]{khashabi2022unifiedqa}
Daniel Khashabi, Yeganeh Kordi, and Hannaneh Hajishirzi. 2022.
\newblock \href {https://arxiv.org/abs/2202.12359} {Unifiedqa-v2: Stronger
  generalization via broader cross-format training}.
\newblock \emph{arXiv preprint arXiv:2202.12359}.

\bibitem[{Khashabi et~al.(2020)Khashabi, Min, Khot, Sabharwal, Tafjord, Clark,
  and Hajishirzi}]{khashabi2020unifiedqa}
Daniel Khashabi, Sewon Min, Tushar Khot, Ashish Sabharwal, Oyvind Tafjord,
  Peter Clark, and Hannaneh Hajishirzi. 2020.
\newblock \href {https://doi.org/10.18653/v1/2020.findings-emnlp.171}
  {{UNIFIEDQA}: Crossing format boundaries with a single {QA} system}.
\newblock In \emph{Findings of the Association for Computational Linguistics:
  EMNLP 2020}, pages 1896--1907, Online. Association for Computational
  Linguistics.

\bibitem[{Khot et~al.(2020)Khot, Clark, Guerquin, Jansen, and
  Sabharwal}]{khot2020qasc}
Tushar Khot, Peter Clark, Michal Guerquin, Peter Jansen, and Ashish Sabharwal.
  2020.
\newblock \href {https://aaai.org/ojs/index.php/AAAI/article/view/6319}
  {{QASC:} {A} dataset for question answering via sentence composition}.
\newblock In \emph{The Thirty-Fourth {AAAI} Conference on Artificial
  Intelligence, {AAAI} 2020, The Thirty-Second Innovative Applications of
  Artificial Intelligence Conference, {IAAI} 2020, The Tenth {AAAI} Symposium
  on Educational Advances in Artificial Intelligence, {EAAI} 2020, New York,
  NY, USA, February 7-12, 2020}, pages 8082--8090. {AAAI} Press.

\bibitem[{Latcinnik and Berant(2020)}]{latcinnik2020explaining}
Veronica Latcinnik and Jonathan Berant. 2020.
\newblock \href {https://arxiv.org/abs/2004.05569} {Explaining question
  answering models through text generation}.
\newblock \emph{arXiv preprint arXiv:2004.05569}.

\bibitem[{Lin et~al.(2020)Lin, Lee, Khanna, and Ren}]{lin2020birds}
Bill~Yuchen Lin, Seyeon Lee, Rahul Khanna, and Xiang Ren. 2020.
\newblock \href {https://doi.org/10.18653/v1/2020.emnlp-main.557} {{B}irds have
  four legs?! {N}umer{S}ense: {P}robing {N}umerical {C}ommonsense {K}nowledge
  of {P}re-{T}rained {L}anguage {M}odels}.
\newblock In \emph{Proceedings of the 2020 Conference on Empirical Methods in
  Natural Language Processing (EMNLP)}, pages 6862--6868, Online. Association
  for Computational Linguistics.

\bibitem[{Lin et~al.(2021)Lin, Wu, Yang, Lee, and Ren}]{lin2021riddlesense}
Bill~Yuchen Lin, Ziyi Wu, Yichi Yang, Dong-Ho Lee, and Xiang Ren. 2021.
\newblock \href {https://doi.org/10.18653/v1/2021.findings-acl.131}
  {{R}iddle{S}ense: Reasoning about riddle questions featuring linguistic
  creativity and commonsense knowledge}.
\newblock In \emph{Findings of the Association for Computational Linguistics:
  ACL-IJCNLP 2021}, pages 1504--1515, Online. Association for Computational
  Linguistics.

\bibitem[{Liu et~al.(2022)Liu, Liu, Lu, Welleck, West, Le~Bras, Choi, and
  Hajishirzi}]{liu2021generated}
Jiacheng Liu, Alisa Liu, Ximing Lu, Sean Welleck, Peter West, Ronan Le~Bras,
  Yejin Choi, and Hannaneh Hajishirzi. 2022.
\newblock \href {https://doi.org/10.18653/v1/2022.acl-long.225} {Generated
  knowledge prompting for commonsense reasoning}.
\newblock In \emph{Proceedings of the 60th Annual Meeting of the Association
  for Computational Linguistics (Volume 1: Long Papers)}, pages 3154--3169,
  Dublin, Ireland. Association for Computational Linguistics.

\bibitem[{Lourie et~al.(2021)Lourie, Le~Bras, Bhagavatula, and
  Choi}]{lourie2021unicorn}
Nicholas Lourie, Ronan Le~Bras, Chandra Bhagavatula, and Yejin Choi. 2021.
\newblock Unicorn on rainbow: A universal commonsense reasoning model on a new
  multitask benchmark.
\newblock In \emph{Proceedings of the AAAI Conference on Artificial
  Intelligence}, volume~35, pages 13480--13488.

\bibitem[{Lu et~al.(2022)Lu, Welleck, Jiang, Hessel, Qin, West, Ammanabrolu,
  and Choi}]{lu2022quark}
Ximing Lu, Sean Welleck, Liwei Jiang, Jack Hessel, Lianhui Qin, Peter West,
  Prithviraj Ammanabrolu, and Yejin Choi. 2022.
\newblock \href {https://arxiv.org/abs/2205.13636} {Quark: Controllable text
  generation with reinforced unlearning}.
\newblock \emph{arXiv preprint arXiv:2205.13636}.

\bibitem[{Mercier and Sperber(2017)}]{Mercier2017TheEO}
Hugo Mercier and Dan Sperber. 2017.
\newblock The enigma of reason.

\bibitem[{Mihaylov et~al.(2018)Mihaylov, Clark, Khot, and
  Sabharwal}]{mihaylov2018can}
Todor Mihaylov, Peter Clark, Tushar Khot, and Ashish Sabharwal. 2018.
\newblock \href {https://doi.org/10.18653/v1/D18-1260} {Can a suit of armor
  conduct electricity? a new dataset for open book question answering}.
\newblock In \emph{Proceedings of the 2018 Conference on Empirical Methods in
  Natural Language Processing}, pages 2381--2391, Brussels, Belgium.
  Association for Computational Linguistics.

\bibitem[{Mitra et~al.(2019)Mitra, Banerjee, Pal, Mishra, and
  Baral}]{mitra2019additional}
Arindam Mitra, Pratyay Banerjee, Kuntal~Kumar Pal, Swaroop Mishra, and Chitta
  Baral. 2019.
\newblock \href {https://arxiv.org/abs/1909.08855} {How additional knowledge
  can improve natural language commonsense question answering?}
\newblock \emph{arXiv preprint arXiv:1909.08855}.

\bibitem[{Nakano et~al.(2021)Nakano, Hilton, Balaji, Wu, Ouyang, Kim, Hesse,
  Jain, Kosaraju, Saunders et~al.}]{nakano2021webgpt}
Reiichiro Nakano, Jacob Hilton, Suchir Balaji, Jeff Wu, Long Ouyang, Christina
  Kim, Christopher Hesse, Shantanu Jain, Vineet Kosaraju, William Saunders,
  et~al. 2021.
\newblock \href {https://arxiv.org/abs/2112.09332} {Webgpt: Browser-assisted
  question-answering with human feedback}.
\newblock \emph{arXiv preprint arXiv:2112.09332}.

\bibitem[{Ouyang et~al.(2022)Ouyang, Wu, Jiang, Almeida, Wainwright, Mishkin,
  Zhang, Agarwal, Slama, Ray et~al.}]{ouyang2022training}
Long Ouyang, Jeff Wu, Xu~Jiang, Diogo Almeida, Carroll~L Wainwright, Pamela
  Mishkin, Chong Zhang, Sandhini Agarwal, Katarina Slama, Alex Ray, et~al.
  2022.
\newblock \href {https://arxiv.org/abs/2203.02155} {Training language models to
  follow instructions with human feedback}.
\newblock \emph{arXiv preprint arXiv:2203.02155}.

\bibitem[{Paranjape et~al.(2021)Paranjape, Michael, Ghazvininejad, Hajishirzi,
  and Zettlemoyer}]{paranjape2021prompting}
Bhargavi Paranjape, Julian Michael, Marjan Ghazvininejad, Hannaneh Hajishirzi,
  and Luke Zettlemoyer. 2021.
\newblock \href {https://doi.org/10.18653/v1/2021.findings-acl.366} {Prompting
  contrastive explanations for commonsense reasoning tasks}.
\newblock In \emph{Findings of the Association for Computational Linguistics:
  ACL-IJCNLP 2021}, pages 4179--4192, Online. Association for Computational
  Linguistics.

\bibitem[{Paulus et~al.(2018)Paulus, Xiong, and Socher}]{paulus2017deep}
Romain Paulus, Caiming Xiong, and Richard Socher. 2018.
\newblock \href {https://openreview.net/forum?id=HkAClQgA-} {A deep reinforced
  model for abstractive summarization}.
\newblock In \emph{6th International Conference on Learning Representations,
  {ICLR} 2018, Vancouver, BC, Canada, April 30 - May 3, 2018, Conference Track
  Proceedings}. OpenReview.net.

\bibitem[{Raffel et~al.(2019)Raffel, Shazeer, Roberts, Lee, Narang, Matena,
  Zhou, Li, and Liu}]{raffel2019exploring}
Colin Raffel, Noam Shazeer, Adam Roberts, Katherine Lee, Sharan Narang, Michael
  Matena, Yanqi Zhou, Wei Li, and Peter~J Liu. 2019.
\newblock \href {https://arxiv.org/abs/1910.10683} {Exploring the limits of
  transfer learning with a unified text-to-text transformer}.
\newblock \emph{arXiv preprint arXiv:1910.10683}.

\bibitem[{Rajani et~al.(2019)Rajani, McCann, Xiong, and
  Socher}]{rajani2019explain}
Nazneen~Fatema Rajani, Bryan McCann, Caiming Xiong, and Richard Socher. 2019.
\newblock \href {https://doi.org/10.18653/v1/P19-1487} {Explain yourself!
  leveraging language models for commonsense reasoning}.
\newblock In \emph{Proceedings of the 57th Annual Meeting of the Association
  for Computational Linguistics}, pages 4932--4942, Florence, Italy.
  Association for Computational Linguistics.

\bibitem[{Sakaguchi et~al.(2021)Sakaguchi, Bras, Bhagavatula, and
  Choi}]{sakaguchi2021winogrande}
Keisuke Sakaguchi, Ronan~Le Bras, Chandra Bhagavatula, and Yejin Choi. 2021.
\newblock Winogrande: An adversarial winograd schema challenge at scale.
\newblock \emph{Communications of the ACM}, 64(9):99--106.

\bibitem[{Sap et~al.(2019)Sap, Rashkin, Chen, Le~Bras, and
  Choi}]{sap2019socialiqa}
Maarten Sap, Hannah Rashkin, Derek Chen, Ronan Le~Bras, and Yejin Choi. 2019.
\newblock \href {https://doi.org/10.18653/v1/D19-1454} {Social {IQ}a:
  Commonsense reasoning about social interactions}.
\newblock In \emph{Proceedings of the 2019 Conference on Empirical Methods in
  Natural Language Processing and the 9th International Joint Conference on
  Natural Language Processing (EMNLP-IJCNLP)}, pages 4463--4473, Hong Kong,
  China. Association for Computational Linguistics.

\bibitem[{Schulman et~al.(2017)Schulman, Wolski, Dhariwal, Radford, and
  Klimov}]{schulman2017proximal}
John Schulman, Filip Wolski, Prafulla Dhariwal, Alec Radford, and Oleg Klimov.
  2017.
\newblock \href {https://arxiv.org/abs/1707.06347} {Proximal policy
  optimization algorithms}.
\newblock \emph{arXiv preprint arXiv:1707.06347}.

\bibitem[{Shen et~al.(2016)Shen, Cheng, He, He, Wu, Sun, and
  Liu}]{shen2015minimum}
Shiqi Shen, Yong Cheng, Zhongjun He, Wei He, Hua Wu, Maosong Sun, and Yang Liu.
  2016.
\newblock \href {https://doi.org/10.18653/v1/P16-1159} {Minimum risk training
  for neural machine translation}.
\newblock In \emph{Proceedings of the 54th Annual Meeting of the Association
  for Computational Linguistics (Volume 1: Long Papers)}, pages 1683--1692,
  Berlin, Germany. Association for Computational Linguistics.

\bibitem[{Shwartz et~al.(2020)Shwartz, West, Le~Bras, Bhagavatula, and
  Choi}]{shwartz2020unsupervised}
Vered Shwartz, Peter West, Ronan Le~Bras, Chandra Bhagavatula, and Yejin Choi.
  2020.
\newblock \href {https://doi.org/10.18653/v1/2020.emnlp-main.373} {Unsupervised
  commonsense question answering with self-talk}.
\newblock In \emph{Proceedings of the 2020 Conference on Empirical Methods in
  Natural Language Processing (EMNLP)}, pages 4615--4629, Online. Association
  for Computational Linguistics.

\bibitem[{Stiennon et~al.(2020)Stiennon, Ouyang, Wu, Ziegler, Lowe, Voss,
  Radford, Amodei, and Christiano}]{stiennon2020learning}
Nisan Stiennon, Long Ouyang, Jeffrey Wu, Daniel Ziegler, Ryan Lowe, Chelsea
  Voss, Alec Radford, Dario Amodei, and Paul~F Christiano. 2020.
\newblock Learning to summarize with human feedback.
\newblock \emph{Advances in Neural Information Processing Systems},
  33:3008--3021.

\bibitem[{Tafjord et~al.(2019)Tafjord, Gardner, Lin, and
  Clark}]{tafjord2019quartz}
Oyvind Tafjord, Matt Gardner, Kevin Lin, and Peter Clark. 2019.
\newblock \href {https://doi.org/10.18653/v1/D19-1608} {{Q}ua{RT}z: An
  open-domain dataset of qualitative relationship questions}.
\newblock In \emph{Proceedings of the 2019 Conference on Empirical Methods in
  Natural Language Processing and the 9th International Joint Conference on
  Natural Language Processing (EMNLP-IJCNLP)}, pages 5941--5946, Hong Kong,
  China. Association for Computational Linguistics.

\bibitem[{Talmor et~al.(2019)Talmor, Herzig, Lourie, and
  Berant}]{talmor2018commonsenseqa}
Alon Talmor, Jonathan Herzig, Nicholas Lourie, and Jonathan Berant. 2019.
\newblock \href {https://doi.org/10.18653/v1/N19-1421} {{C}ommonsense{QA}: A
  question answering challenge targeting commonsense knowledge}.
\newblock In \emph{Proceedings of the 2019 Conference of the North {A}merican
  Chapter of the Association for Computational Linguistics: Human Language
  Technologies, Volume 1 (Long and Short Papers)}, pages 4149--4158,
  Minneapolis, Minnesota. Association for Computational Linguistics.

\bibitem[{Wang et~al.(2022)Wang, Srikumar, Hajishirzi, and
  Smith}]{wang2022elaboration}
Wenya Wang, Vivek Srikumar, Hanna Hajishirzi, and Noah~A Smith. 2022.
\newblock Elaboration-generating commonsense question answering at scale.
\newblock \emph{arXiv preprint arXiv:2209.01232}.

\bibitem[{Wei et~al.(2022)Wei, Wang, Schuurmans, Bosma, Chi, Le, and
  Zhou}]{wei2022chain}
Jason Wei, Xuezhi Wang, Dale Schuurmans, Maarten Bosma, Ed~Chi, Quoc Le, and
  Denny Zhou. 2022.
\newblock \href {https://arxiv.org/abs/2201.11903} {Chain of thought prompting
  elicits reasoning in large language models}.
\newblock \emph{arXiv preprint arXiv:2201.11903}.

\bibitem[{Wu et~al.(2016)Wu, Schuster, Chen, Le, Norouzi, Macherey, Krikun,
  Cao, Gao, Macherey et~al.}]{wu2016google}
Yonghui Wu, Mike Schuster, Zhifeng Chen, Quoc~V Le, Mohammad Norouzi, Wolfgang
  Macherey, Maxim Krikun, Yuan Cao, Qin Gao, Klaus Macherey, et~al. 2016.
\newblock \href {https://arxiv.org/abs/1609.08144} {Google's neural machine
  translation system: Bridging the gap between human and machine translation}.
\newblock \emph{arXiv preprint arXiv:1609.08144}.

\bibitem[{Zellers et~al.(2019)Zellers, Holtzman, Bisk, Farhadi, and
  Choi}]{zellers2019hellaswag}
Rowan Zellers, Ari Holtzman, Yonatan Bisk, Ali Farhadi, and Yejin Choi. 2019.
\newblock \href {https://doi.org/10.18653/v1/P19-1472} {{H}ella{S}wag: Can a
  machine really finish your sentence?}
\newblock In \emph{Proceedings of the 57th Annual Meeting of the Association
  for Computational Linguistics}, pages 4791--4800, Florence, Italy.
  Association for Computational Linguistics.

\end{thebibliography}

\clearpage
\appendix

\section{Additional Experimental Details}

\subsection{Hyperparameters}
\label{sec:hypers}

See \autoref{tab:hypers}.

\subsection{Baselines}
\label{sec:baselines}

\begin{table*}[b]
\small
\centering
\begin{tabular}{l l}
\toprule
\textbf{Question Template} & \textbf{Answer Template} \\
\midrule
What is the definition of & The definition of \_ is \\
What is the main purpose of & The purpose of \_ is to \\
What is the main function of a & The main function of a \_ is \\
What are the properties of a & The properties of a \_ are that \\
What is a & \_ is \\
What happened as a result of & As a result of \_, \\
What might have caused & The cause of \_ was \\
What is a part of & A part of \_ is \\
What is an example of & An example of \_ is \\
How would you & One would \_ by \\
\bottomrule
\end{tabular}
\caption{Templates used in the self-talk baseline.}
\label{tab:selftalk_templates}
\end{table*}

\paragraph{Self-talk.}
We generate $M = 10$ knowledge per question with GPT-3-Curie, using 10 pairs of question-answer templates adapted from \citet{shwartz2020unsupervised}.
We generate one knowledge from each template: first, query GPT-3 with the question template and using nucleus sampling ($p = 0.2$) to obtain a full question; next, query GPT-3 again with both the full question and the corresponding answer template, this time using nucleus sampling ($p = 0.5$), to obtain a full answer sentence.
The answer sentence will be treated as the knowledge.

\paragraph{DREAM.}
We generate $M = 10$ scene elaborations per question with the DREAM (11B) model.
\citet{gu2021dream} proposes 4 types of scene elaborations: motivation ($M$), emotion ($E$), rule-of-thumb ($ROT$), and consequence ($Con$).
Each is associated with a control code that guides the DREAM model.
We generate 2 or 3 scene elaborations for each type, making a total of 10 per question.

\newpage

\section{Additional Analysis}
\label{sec:analysis_more}

\autoref{tab:qual_sematic} through \ref{tab:qual_ethics} show more analysis of knowledge generated by \methodname{}.
\autoref{tab:qual_sematic} shows semantically problematic knowledge.
\autoref{tab:qual_social} shows knowledge that express some social value.
\autoref{tab:qual_culture} shows knowledge that are culture-specific.
\autoref{tab:qual_ethics} shows knowledge that have potential ethical risks.
All examples are taken from the validation set of the respective dataset.

\section{Prompts for Getting Silver Knowledge from GPT-3}
\label{sec:prompts}

See \autoref{tab:prompt_obqa} through \ref{tab:prompt_wg}.

\begin{table*}[t]
\small
\centering
\begin{tabular}{c c l}
\toprule
\textbf{Symbol} & \textbf{Value} & \textbf{Description} \\
\midrule
\multicolumn{3}{c}{\textsc{Getting Silver Knowledge from Few-Shot GPT-3}} \\
\midrule
$M$ & 20 & Number of knowledge statements to sample from GPT-3, per question. \\
$p$ & 0.5 & Parameter for nucleus sampling from GPT-3. \\
$L_{\text{output}}$ & 64 & Max length of output from GPT-3. \\
\midrule
\multicolumn{3}{c}{\textsc{Stage I: Imitation Learning}} \\
\midrule
$L_{\text{input}}$ & 256 & Max length of input to \methodname{} (i.e. question plus choices). \\
$L_{\text{output}}$ & 64 & Max length of output from \methodname{} (i.e. generated knowledge). \\
$B$ & 64 & Batch size for training. \\
$S$ & 50,000 & Total number of training steps. \\
$\eta$ & $1 \times 10^{-5}$ & Learning rate of Adam optimizer. \\
\midrule
\multicolumn{3}{c}{\textsc{Stage II: Reinforcement Learning}} \\
\midrule
$\alpha$ & 1.0 & Weight of value model loss in PPO. \\
$\beta$ & 0.2 & Weight of entropy bonus term in reward. \\
$\gamma$ & 1.0 & Discount factor for rewards. \\
$\lambda$ & 0.95 & Parameter for advantage estimation. \\
$\eps$ & 0.2 & Clipping range for the \textit{clipped surrogate objective}. \\
$L_{\text{input}}$ & 256 & Max length of input to \methodname{} (i.e. question plus choices). \\
$L_{\text{output}}$ & 32 & Max length of output from \methodname{} (i.e. generated knowledge). \\
$\tau$ & 0.7 & Temperature for knowledge sampling in PPO training. \\
$E$ & 1M & Total number of training episodes. \\
$B$ & 64 & Batch size for training. \\
$S$ & 15,625 & Total number of training steps. \\
$s$ & 4 & Interval (in steps) for updating the lagging models (policy and value). \\
$\eta$ & $2 \times 10^{-5}$ & Learning rate of Adam optimizer (with a linear learning rate decay schedule). \\
\midrule
\multicolumn{3}{c}{\textsc{Inference}} \\
\midrule
$M$ & 10 & Number of knowledge statements to sample from \methodname{}, per question. \\
$p$ & 0.5 & Parameter for nucleus sampling from \methodname{}. \\
$L_{\text{input}}$ & 256 & Max length of input to \methodname{} (i.e. question plus choices). \\
$L_{\text{output}}$ & 32 & Max length of output from \methodname{}. \\
\bottomrule
\end{tabular}
\caption{Hyperparameter settings.}
\label{tab:hypers}
\end{table*}

\begin{table*}[t]
\setlength{\tabcolsep}{3pt}
\centering
\resizebox{0.6\textwidth}{!}{%
\begin{tabular}{l c cc cc}
\toprule
\textbf{Dataset} $\rightarrow$ & \multirow{2}{*}{OBQA} & \multicolumn{2}{c}{ARC} & \multicolumn{2}{c}{AI2Science} \\
\textbf{Method} $\downarrow$ &  & \small{easy} & \small{hard} & \small{elem} & \small{mid} \\
\midrule
UQA-large (0.77B) & 70.20 & 69.12 & 55.85 & 69.11 & 64.80 \\
\quad + Few-shot GPT-3-Curie (13B) & 68.80 & 71.05 & 56.52 & 70.73 & 65.60 \\
\midrule
\quad \textbf{+ \methodname{}-large (0.77B) [ours]} & 69.60 & 67.72 & 55.18 & 68.29 & 63.20 \\
\bottomrule
\end{tabular}
}%
\vspace{-4pt}
\caption{
    Results on the other 3 \textbf{seen} datasets.
    All experiments use UnifiedQA-large as the QA model, and optionally uses knowledge from one of the knowledge generation models.
    On these datasets, \methodname{}-generated knowledge does not give an improvement over the vanilla QA baseline.
}
\label{tab:results_id_nope}
\end{table*}

\begin{table*}[h!]
\setlength{\tabcolsep}{3pt}
\footnotesize
\centering
\begin{tabular}{cp{200pt} p{144pt}}
\toprule
\textbf{Task} & Question / \textbf{Knowledge} & \textbf{Problem with the knowledge} \\
\midrule
RS & I am a fruit. I am tasty and provide lots of energy. You can also find me in a calendar. What am I? \newline (A) shop (B) choppers \Right{(C) date} (D) european (E) movie \newline \textbf{The fruit of the year is the date.} & \textbf{Ambiguous} \newline The knowledge does not specify which year, so it is not possible to verify its factuality. \\
\midrule
QuaRTz & Carla is pushing two carts down a street. One cart is heavy, the other is light. The heavy cart, compared to the light one, will accelerate \newline \Right{(A) slowly} (B) quickly \newline \textbf{The heavy cart will accelerate slower than the light one.} & \textbf{Under-specified} \newline Missing the control factor that the forces exerted on the carts are the same. Though this flaw also exists in the question itself. \\
\midrule
RS & What has a head at night but not in the morning? \newline (A) sleeping \Right{(B) pillow} (C) neck (D) shampoo (E) beer \newline \textbf{Sleeping animals have a head at night.} & \textbf{Over-specified} \newline It would be normal to say "animals have a head". It is weird to qualify with "sleeping" and "at night". \\
\midrule
WG & Because William developed a sore knee during his race against against Kyle, \_ won the race. \newline (A) William \Right{(B) Kyle} \newline \textbf{William was not able to run the race.} & \textbf{Over-confident} \newline William might still be able to finish the race with a sore knee. \\
\midrule
RS & what is the begining of enternity the end of life the end of time and the end to every race \newline \Right{(A) e} (B) quick (C) finality (D) fix (E) habit \newline \textbf{The end of every race is the end of every race.} & \textbf{Tautological} \newline This knowledge does not give any meaningful information. \\
\midrule
QuaRTz & Sharon is conducting an experiment on valence electrons and soon discovers that when they are closer to the nucleus, they are \_\_\_\_\_ easily removed from the atom. \newline (A) more \Right{(B) less} \newline \textbf{Valence electrons are more prone to being removed from the atom.} & \textbf{Not applicable} \newline This knowledge implicitly compares the removability of valence vs. non-valence electrons. However, the question needs a comparison of valence electrons in atoms of different sizes, so the knowledge cannot be applied to answering this question. \\
\bottomrule
\end{tabular}
\caption{
    Examples of knowledge generated by \methodname{} which are semantically problematic.
}
\label{tab:qual_sematic}
\end{table*}
\newpage
\begin{table*}[h!]
\setlength{\tabcolsep}{3pt}
\footnotesize
\centering
\begin{tabular}{cp{200pt} p{144pt}}
\toprule
\textbf{Task} & Question / \textbf{Knowledge} & \textbf{Problem with the knowledge} \\
\midrule
SIQA & Riley broke loose from the house. He thought he would never get out of there. Why did Riley do this? \newline (A) Stay in the house longer \Right{(B) think about his life} (C) go home for a while \newline \textbf{Breaking out of a bad habit is usually a bad idea.} & \textbf{Social value} \newline This knowledge is a generally true statement, so we labeled it as factual. \\
\midrule
SIQA & Tracy heard a faint buzzing noise and immediately ran for her life. How would you describe Tracy? \newline \Right{(A) scared of bees} (B) sad (C) not phased by bees \newline \textbf{One should not be scared of bees.} & \textbf{Social value} \newline It is hard to decide whether this knowledge should be considered factual or not. \\
\midrule
SIQA & Remy gave Skylar's Netflix account password to one of Remy's other friends. How would Skylar feel as a result? \newline (A) like a bad friend (B) excited \Right{(C) used} \newline \textbf{A friend can be used by a friend.} & \textbf{Social value} \newline It is ambiguous whether \textit{can} means \textit{it is possible that ...}, or \textit{ought to}. If it is the latter, then the knowledge is promoting some problematic social value. \\
\midrule
SIQA & Riley was the best of friends with the boy with cancer. What will Riley want to do next? \newline \Right{(A) visit the hospital} (B) shun the friend (C) become friends with the boy with cancer too \newline \textbf{One should visit their sick friend.} & \textbf{Social value} \newline It is generally a kind thing to visit a sick friend. However, it is conceivable that the friend needs to recover in peace or has some infectious disease, which renders a visit inappropriate. \\
\midrule
SIQA & Carson tried to fight Robin last night because Robin hurt Carson a lot. What will Carson want to do next? \newline (A) apologize (B) do nothing \Right{(C) hurt Robin} \newline \textbf{One should apologize when they hurt someone.} & \textbf{Social value} \newline This knowledge is generally accepted. However, there are extenuating circumstances where hurting someone does not need an apology (e.g. hurting a violent criminal to protect oneself). \\
\midrule
SIQA & Bailey told Alex to send the pdf because they didn't want to do it themselves. How would Alex feel as a result? \newline (A) lazy about work (B) happy \Right{(C) angry} \newline \textbf{One should be willing to help others.} & \textbf{Social value} \newline This knowledge is generally accepted, but it is not a good fit to the question's context. It is normal to be emotional when being ordered to do something on other's behalf. \\
\midrule
SIQA & Kendall wrapped a bandage around my neck after getting injured in a fight. What will Kendall want to do next? \newline (A) harm them (B) punish them \Right{(C) protect them} \newline \textbf{One should help others in need.} & \textbf{Social value} \newline This knowledge is generally accepted, and appropriate to the question's context. \\
\bottomrule
\end{tabular}
\caption{
    Examples of knowledge generated by \methodname{} that express some social value.
}
\label{tab:qual_social}
\end{table*}
\newpage
\begin{table*}[h!]
\setlength{\tabcolsep}{3pt}
\footnotesize
\centering
\begin{tabular}{cp{200pt} p{144pt}}
\toprule
\textbf{Task} & Question / \textbf{Knowledge} & \textbf{Problem with the knowledge} \\
\midrule
SIQA & Remy made hay getting home from school on Friday the 13th. Why did Remy do this? \newline (A) go to school before this (B) had heard that on Friday the 13th, God would bless you if you just went home and hid \Right{(C) had heard that on Friday the 13th, that people pranked other people really bad} \newline \textbf{People are more likely to be pranked on Friday the 13th.} & \textbf{Culture-specific} \newline This knowledge largely applies within western, especially Christian, culture. \\
\midrule
WG & Lindsey like to read graphic novels but Natalie liked classic literature to read. \_ bought the new Frank Miller comic at the book store. \newline \Right{(A) Lindsey} (B) Natalie \newline \textbf{Graphic novels are popular among teens.} & \textbf{Culture-specific} \newline This knowledge could be correct or not depending on the culture or geographical location. \\
\midrule
CSQA & Where would a person live that isn't in the metro area but still has good schools? \newline (A) woods (B) bed \Right{(C) suburbs} (D) rural (E) neighborhood \newline \textbf{A person who lives in the suburbs is likely to have good schools.} & \textbf{Culture-specific} \newline This knowledge could be correct or not depending on the culture or geographical location, and it is not even always true within the U.S. \\
\midrule
NS & most people in the u.s. works <mask> consecutive hours in a business day. \newline (A) no (B) one (C) two (D) three (E) four (F) five (G) six (H) seven \Right{(I) eight} (J) nine (K) ten \newline \textbf{Most people work eight hours in a day.} & \textbf{Culture-specific} \newline This knowledge is an over-generalization. In some other cultures, most people work more then eight hours, or less. \\
\midrule
CSQA & What will god never do according to religion? \newline (A) anything \Right{(B) judge people} (C) work miracles (D) judge men (E) everywhere \newline \textbf{God is a judge of people.} & \textbf{Culture-specific} \newline This is a topic specific to the religion. \\
\bottomrule
\end{tabular}
\caption{
    Examples of knowledge generated by \methodname{} that are culture-specific.
}
\label{tab:qual_culture}
\end{table*}
\begin{table*}[h!]
\setlength{\tabcolsep}{3pt}
\footnotesize
\centering
\begin{tabular}{cp{200pt} p{144pt}}
\toprule
\textbf{Task} & Question / \textbf{Knowledge} & \textbf{Problem with the knowledge} \\
\midrule
RS & A woman screams. You gointo a shop and find a man lying dead on the floor. There are three people in the room. A postman, a milkman and a doctor. Who killed him? \newline \Right{(A) doctor} (B) police station (C) liars (D) radio (E) water fountain \newline \textbf{The doctor is the one who killed the man.} & \textbf{Ethical risk} \newline This knowledge might impress bias towards a certain profession. However, the question itself should probably take the blame, because unfortunately, "doctor" is the only valid choice given the question's context. \\
\midrule
PIQA & crossing out \newline (A) ugly people from the bachelor is cool \Right{(B) ugly people from planet earth is cool} \newline \textbf{Crossing out ugly people from planet earth is cool.} & \textbf{Ethical risk} \newline This knowledge might impress bias towards a group with certain physical characteristic. \\
\midrule
CSQA & With the card slot lit up he knew how to get started finding his balance with what? \newline (A) slot machine (B) ticket machine (C) bank machine (D) telephone \Right{(E) automated teller} \newline \textbf{A slot machine is a machine that takes cards and uses them to make money.} & \textbf{Ethical risk} \newline This knowledge might advocate for gambling. \\
\bottomrule
\end{tabular}
\caption{
    Examples of knowledge generated by \methodname{} that have potential ethical risks.
}
\label{tab:qual_ethics}
\end{table*}

\begin{table*}[h!]
\footnotesize
\centering
\begin{tabular}{r p{12cm}}
\textbf{Task} & \textbf{Prompt} \\
\midrule
OBQA & Input: The sun is responsible for \textbackslash{}n (A) puppies learning new tricks (B) children growing up and getting old (C) flowers wilting in a vase (D) plants sprouting, blooming and wilting \\
& Knowledge: \textbf{Natural light provides energy for photosynthesis.} \\
\\
& Input: Poison causes harm to which of the following? \textbackslash{}n (A) a Tree (B) a robot (C) a house (D) a car \\
& Knowledge: \textbf{Living organisms are susceptible to poisonous matter.} \\
\\
& Input: As a car approaches you in the night \textbackslash{}n (A) the headlights become more intense (B) the headlights recede into the dark (C) the headlights remain at a constant (D) the headlights turn off \\
& Knowledge: \textbf{The intensity of light increases when observed from a shorter distance.} \\
\\
& Input: When the weather changes as it does from Christmas to Easter, \textbackslash{}n (A) the air may chill (B) the ground may freeze (C) the plants may die (D) the ground may warm \\
& Knowledge: \textbf{Christmas is in winter and Easter is in spring.} \\
\\
& Input: Using mirrors to focus collected light from heavenly bodies allows \textbackslash{}n (A) detailed observation (B) foregone conclusions (C) radiation experiments (D) celestial music \\
& Knowledge: \textbf{Telescopes use mirrors to focus light from the stars.} \\
\\
& Input: \{question\}\\
& Knowledge:
\end{tabular}
\caption{
    Prompt for OpenBookQA.
}
\label{tab:prompt_obqa}
\end{table*}
\begin{table*}[h!]
\footnotesize
\centering
\begin{tabular}{r p{12cm}}
\textbf{Task} & \textbf{Prompt} \\
\midrule
ARC & Input: George wants to warm his hands quickly by rubbing them. Which skin surface will produce the most heat? \textbackslash{}n (A) dry palms (B) wet palms (C) palms covered with oil (D) palms covered with lotion \\
& Knowledge: \textbf{Rubbing hands produces heat because of friction.} \\
\\
& Input: Which of the following is an example of a physical change? \textbackslash{}n (A) lighting a match (B) breaking a glass (C) burning of gasoline (D) rusting of iron \\
& Knowledge: \textbf{Physical changes must not involve chemical changes such as combustion and rusting.} \\
\\
& Input: On Earth, water can be a solid, a liquid, or a gas. Which energy source has the greatest influence on the state of matter of water? \textbackslash{}n (A) the sun (B) the wind (C) ocean currents (D) the metal core \\
& Knowledge: \textbf{Earth's water circulation is mostly driven by heat radiated from the sun.} \\
\\
& Input: What do cells break down to produce energy? \textbackslash{}n (A) food (B) water (C) chlorophyll (D) carbon dioxide \\
& Knowledge: \textbf{Food contain calories.} \\
\\
& Input: What characteristic of DNA results in cell differentiation in developing embryos? \textbackslash{}n (A) which genes are present (B) how many copies of each gene are present (C) which genes are active (D) what protein is produced by a gene \\
& Knowledge: \textbf{Cell differentiation is caused by selective expression of genes.} \\
\\
& Input: \{question\}\\
& Knowledge:
\end{tabular}
\caption{
    Prompt for ARC.
}
\label{tab:prompt_arc}
\end{table*}
\begin{table*}[h!]
\footnotesize
\centering
\begin{tabular}{r p{12cm}}
\textbf{Task} & \textbf{Prompt} \\
\midrule
AI2Sci & Input: Which is a nonrenewable natural resource that is used to make electrical energy? \textbackslash{}n (A) coal (B) wind (C) water (D) thermal \\
& Knowledge: \textbf{Fossil fuel is nonrenewable natural resource.} \\
\\
& Input: Which adaptation will warn predators not to eat an animal? \textbackslash{}n (A) bright colors (B) bulging eyes (C) geometric shapes (D) poisonous secretions \\
& Knowledge: \textbf{Bright colors in animals are usually a sign of being poisonous.} \\
\\
& Input: An Italian scientist named Alessandro Volta invented the Voltaic pile in 1800. It was able to produce a steady electrical current. Based on this description, what is the modern equivalent of the Voltaic pile? \textbackslash{}n (A) a wire (B) a battery (C) a resistor (D) a light bulb \\
& Knowledge: \textbf{Batteries can produce steady electrical current.} \\
\\
& Input: What is the best measure to use in determining the effect of solar energy on Earth's atmosphere? \textbackslash{}n (A) the temperature of the air (B) the temperature of the ocean (C) the density of clouds in the sky (D) the amount of rainfall on a rainy day \\
& Knowledge: \textbf{Solar radiation converts to heat in Earth's atmosphere.} \\
\\
& Input: Which nongaseous compound can be made from two elements that are gases at room temperature? \textbackslash{}n (A) water (B) table salt (C) iron oxide (D) carbon dioxide \\
& Knowledge: \textbf{Water molecules are made of Hydrogen and Oxygen.} \\
\\
& Input: \{question\}\\
& Knowledge:
\end{tabular}
\caption{
    Prompt for AI2Science.
}
\label{tab:prompt_ai2sci}
\end{table*}
\begin{table*}[h!]
\footnotesize
\centering
\begin{tabular}{r p{12cm}}
\textbf{Task} & \textbf{Prompt} \\
\midrule
CSQA & Input: Google Maps and other highway and street GPS services have replaced what? \textbackslash{}n (A) united states (B) mexico (C) countryside (D) atlas (E) oceans \\
& Knowledge: \textbf{Electronic maps are the modern version of paper atlas.} \\
\\
& Input: The fox walked from the city into the forest, what was it looking for? \textbackslash{}n (A) pretty flowers. (B) hen house (C) natural habitat (D) storybook (E) dense forest \\
& Knowledge: \textbf{Natural habitats are usually away from cities.} \\
\\
& Input: You can share files with someone if you have a connection to a what? \textbackslash{}n (A) freeway (B) radio (C) wires (D) computer network (E) electrical circuit \\
& Knowledge: \textbf{Files can be shared over the Internet.} \\
\\
& Input: Too many people want exotic snakes.  The demand is driving what to carry them? \textbackslash{}n (A) ditch (B) shop (C) north america (D) pet shops (E) outdoors \\
& Knowledge: \textbf{Some people raise snakes as pets.} \\
\\
& Input: The body guard was good at his duties, he made the person who hired him what? \textbackslash{}n (A) better job (B) irritated (C) feel safe (D) save money (E) headache \\
& Knowledge: \textbf{The job of body guards is to ensure the safety and security of the employer.} \\
\\
& Input: \{question\}\\
& Knowledge:
\end{tabular}
\caption{
    Prompt for CommonsenseQA.
}
\label{tab:prompt_csqa}
\end{table*}
\begin{table*}[h!]
\footnotesize
\centering
\begin{tabular}{r p{12cm}}
\textbf{Task} & \textbf{Prompt} \\
\midrule
QASC & Input: What type of water formation is formed by clouds? \textbackslash{}n (A) pearls (B) streams (C) shells (D) diamonds (E) rain (F) beads (G) cooled (H) liquid \\
& Knowledge: \textbf{Clouds are made of water vapor.} \\
\\
& Input: What can prevent food spoilage? \textbackslash{}n (A) prolactin release (B) one celled organisms (C) hydrating food (D) cleaning food (E) airing out food (F) Electric generators (G) a hydraulic system (H) dehydrating food \\
& Knowledge: \textbf{Dehydrating food is used for preserving food.} \\
\\
& Input: The process by which genes are passed is \textbackslash{}n (A) Most plants (B) flow of electrons (C) mitosis (D) Summer (E) respiration (F) mutation (G) mechanical (H) reproduction \\
& Knowledge: \textbf{Genes are passed from parent to offspring.} \\
\\
& Input: The stomach does what in the body? \textbackslash{}n (A) decreases its bodily water (B) kills all germs (C) breaks food into nutrients (D) stores bile (E) heat is produced (F) extracts water from food (G) get chemical reactions started (H) cause people to become sick. \\
& Knowledge: \textbf{The stomach is part of the digestive system.} \\
\\
& Input: What can cause rocks to break down? \textbackslash{}n (A) Wind Barriers (B) Protective Barriers (C) Stone Sealers (D) wind (E) mines (F) Water (G) erosion (H) Gravity \\
& Knowledge: \textbf{Mechanical weathering is when rocks are broken down by mechanical means.} \\
\\
& Input: \{question\}\\
& Knowledge:
\end{tabular}
\caption{
    Prompt for QASC.
}
\label{tab:prompt_qasc}
\end{table*}
\begin{table*}[h!]
\footnotesize
\centering
\begin{tabular}{r p{12cm}}
\textbf{Task} & \textbf{Prompt} \\
\midrule
PIQA & Input: how do you flood a room? \textbackslash{}n (A) fill it with objects. (B) fill it with water. \\
& Knowledge: \textbf{Too much water can cause flooding.} \\
\\
& Input: How can I get oil stains out of my driveway? \textbackslash{}n (A) Douse each stain with a couple cans of beer. (B) Douse each stain with a couple cans of soda. \\
& Knowledge: \textbf{Sodium carbonate solution can wash away oil stains.} \\
\\
& Input: Soothe a painful sunburn. \textbackslash{}n (A) Wait until brewed tea bag is cool, then apply on burn. (B) Wait until brewed tea bag is hot, then apply on burn. \\
& Knowledge: \textbf{Sunburn can be alleviated by applying cold material.} \\
\\
& Input: What can I use for fuel in an alcohol stove? \textbackslash{}n (A) Use acetone. (B) Use vinegar. \\
& Knowledge: \textbf{Acetone is flammable, while vinegar is not.} \\
\\
& Input: How can I cut the handles of metal cutlery? \textbackslash{}n (A) Use a hand saw to cut the handles. (B) Use a hand drill to cut the handles. \\
& Knowledge: \textbf{A hand saw is used for making cuts; a hand drill is used for making holes.} \\
\\
& Input: \{question\}\\
& Knowledge:
\end{tabular}
\caption{
    Prompt for PhysicalIQA.
}
\label{tab:prompt_piqa}
\end{table*}
\begin{table*}[h!]
\footnotesize
\centering
\begin{tabular}{r p{12cm}}
\textbf{Task} & \textbf{Prompt} \\
\midrule
SIQA & Input: What will Quinn want to do next? \textbackslash{}n (A) Eat messy snacks (B) help out a friend (C) Pick up the dirty clothes \textbackslash{}n Quinn wanted to help me clean my room up because it was so messy. \\
& Knowledge: \textbf{A messy room likely contains dirty clothes.} \\
\\
& Input: What will Aubrey want to do next? \textbackslash{}n (A) help Aubrey go back home (B) keep on partying without the mom (C) going on with the mom \textbackslash{}n Sasha's mom passed out in the middle of the party. Aubrey took Sasha's mom to the hospital. \\
& Knowledge: \textbf{One should attend to their sick family member.} \\
\\
& Input: How would Jan feel afterwards? \textbackslash{}n (A) scared of losing the cat (B) normal (C) relieved for fixing the problem \textbackslash{}n Their cat kept trying to escape out of the window, so Jan placed an obstacle in the way. \\
& Knowledge: \textbf{One usually has positive emotions after solving a problem.} \\
\\
& Input: How would Sydney feel afterwards? \textbackslash{}n (A) affected (B) like they released their tension (C) worse \textbackslash{}n Sydney had so much pent up emotion, they burst into tears at work. \\
& Knowledge: \textbf{Crying can be a catharsis.} \\
\\
& Input: What does Sydney need to do before this? \textbackslash{}n (A) be bad at her job (B) do a good job (C) be lazy \textbackslash{}n Sydney got a raise and a new promotion. \\
& Knowledge: \textbf{Pay raise and promotion are usually results of good job performance.} \\
\\
& Input: \{question\}\\
& Knowledge:
\end{tabular}
\caption{
    Prompt for SocialIQA.
}
\label{tab:prompt_siqa}
\end{table*}
\begin{table*}[h!]
\footnotesize
\centering
\begin{tabular}{r p{12cm}}
\textbf{Task} & \textbf{Prompt} \\
\midrule
WG & Input: The GPS and map helped me navigate home.  I got lost when the \_ got turned off. \textbackslash{}n (A) GPS (B) map \\
& Knowledge: \textbf{A GPS device is electronic, while a map is paper-based.} \\
\\
& Input: I picked up a bag of peanuts and raisins for a snack. I wanted a sweeter snack out so I ate the \_ for now. \textbackslash{}n (A) raisins (B) peanuts \\
& Knowledge: \textbf{Peanuts contain a lot of fat. Raisins contain a lot of sugar.} \\
\\
& Input: The geese prefer to nest in the fields rather than the forests because in the \_ predators are more hidden. \textbackslash{}n (A) fields (B) forests \\
& Knowledge: \textbf{There are more trees in the forests than in the fields.} \\
\\
& Input: Once in Poland, Dennis enjoyed the trip more than Jason because \_ had a shallow understanding of the Polish language. \textbackslash{}n (A) Dennis (B) Jason \\
& Knowledge: \textbf{Those who know the native language would enjoy the trip better.} \\
\\
& Input: Adam put handwash only clothes in the washer but Aaron washed them by hand as \_ was lazy. \textbackslash{}n (A) Adam (B) Aaron \\
& Knowledge: \textbf{Washing clothes with washer takes less effort than by hand.} \\
\\
& Input: \{question\}\\
& Knowledge:
\end{tabular}
\caption{
    Prompt for Winogrande.
}
\label{tab:prompt_wg}
\end{table*}

\end{document}